%% file: main_isit2024.tex
\documentclass[conference,letterpaper]{IEEEtran}

\newif\ifcondition 

\conditiontrue

\usepackage[utf8]{inputenc} 
\usepackage[T1]{fontenc}
\usepackage{url}
\usepackage{ifthen}
\usepackage[nocompress]{cite}
\usepackage[cmex10]{amsmath} 

\usepackage{ml_package}
\usepackage{tikz}
\usepackage{standalone}
\usepackage{amssymb,amsfonts}

\newcommand\blfootnote[1]{
  \begingroup
  \renewcommand\thefootnote{}\footnote{#1}%
  \addtocounter{footnote}{-1}%
  \endgroup
}

\begin{document}
\title{Cross-Validation Conformal Risk Control} 

\author{%
  \IEEEauthorblockN{Kfir~M.~Cohen, Sangwoo~Park, Osvaldo~Simeone}
  \IEEEauthorblockA{\small{King’s Communications, Learning, and Information Processing (KCLIP) lab}\\
                    \small{Centre for Intelligent Information Processing Systems (CIIPS)}\\ 
                    \small{Department of Engineering, King’s College London}
                    }
  \and
  \IEEEauthorblockN{Shlomo~Shamai~(Shitz)}
  \IEEEauthorblockA{\small{Viterbi Faculty of Electrical and Computing Engineering}\\ 
                    \small{Technion—Israel Institute of Technology}
                    }
}

\maketitle

\ifcondition
    \pagestyle{plain} 
\fi


\begin{abstract}
   Conformal risk control (CRC) is a recently proposed technique that  applies post-hoc to a conventional point predictor to provide calibration guarantees. Generalizing conformal prediction (CP), with CRC, calibration is ensured for a set predictor that is extracted from the point predictor to control a  risk function such as the probability of miscoverage or the false negative rate.  The original CRC requires the available data set to  be split  between   training and validation data sets. This can be problematic when data availability is limited, resulting in inefficient set predictors.   In this paper,  a novel  CRC method is introduced that is based on cross-validation, rather than on validation as the original CRC. The proposed cross-validation CRC (CV-CRC) extends a version of the jackknife-minmax from CP to CRC, allowing for the control of a broader range of risk functions. CV-CRC is proved to offer theoretical guarantees on the average risk of the set predictor. Furthermore, numerical experiments show that CV-CRC can reduce the average set size with respect to CRC when the available data are limited. 
\end{abstract}


\section{Introduction}

\blfootnote{The work of Kfir M. Cohen, Sangwoo Park and Osvaldo Simeone has been supported by the European Research Council (ERC) under the European Union’s Horizon 2020 research and innovation programme, grant agreement No. 725731. The work of Osvaldo Simeone has also been supported by an Open Fellowship of the EPSRC with reference EP/W024101/1, by the European Union’s Horizon Europe Project CENTRIC under Grant 101096379, and by Project REASON, a UK Government funded project under the Future Open Networks Research Challenge (FONRC) sponsored by the Department of Science Innovation and Technology (DSIT). The work of Shlomo Shamai has been supported by  the German Research Foundation (DFG) via the German-Israeli Project Cooperation (DIP), under Project SH 1937/1-1. The authors acknowledge King's Computational Research, Engineering and Technology Environment (CREATE). Retrieved January 18, 2024, from \url{https://doi.org/10.18742/rnvf-m076}.}
\blfootnote{KMC, SP and OS conceived the project; KMC and SP developed the theory with the supervision and guidance of OS; KMC performed the simulation; KMC, SP and OS prepared the manuscript; and SS reviewed the text and contributed to the vision of the paper. All authors discussed the results and contributed to the final manuscript.}

\subsection{Context and Motivation}
One of the key requirements for the application of artificial intelligence (AI) tools to risk-sensitive fields such as healthcare and engineering is the capacity of AI algorithms to quantify their uncertainty \cite{tran2022plex,rajendran2023towards}.  This requires guarantees on the adherence of the ``error bars''  produced by the AI model to the true predictive uncertainty. The predictive uncertainty encompasses both the epistemic uncertainty caused by limited availability of data and the aleatoric uncertainty inherent in the randomness of data generation \cite{simeone2022machine}. Without making strong assumptions on the data generation mechanism it is generally impossible to provide strict uncertainty quantification guarantees for any input, but  assumption-free guarantees can be established on average over validation and test data \cite{lei2014distribution}. \emph{Conformal prediction} (CP) \cite{vovk2005algorithmic,angelopoulos2023conformal}, and its extension \emph{conformal risk control} (CRC) \cite{angelopoulos2022conformal}, are widely established methodologies for the evaluation of predictors with provable uncertainty quantification properties.

\begin{figure}
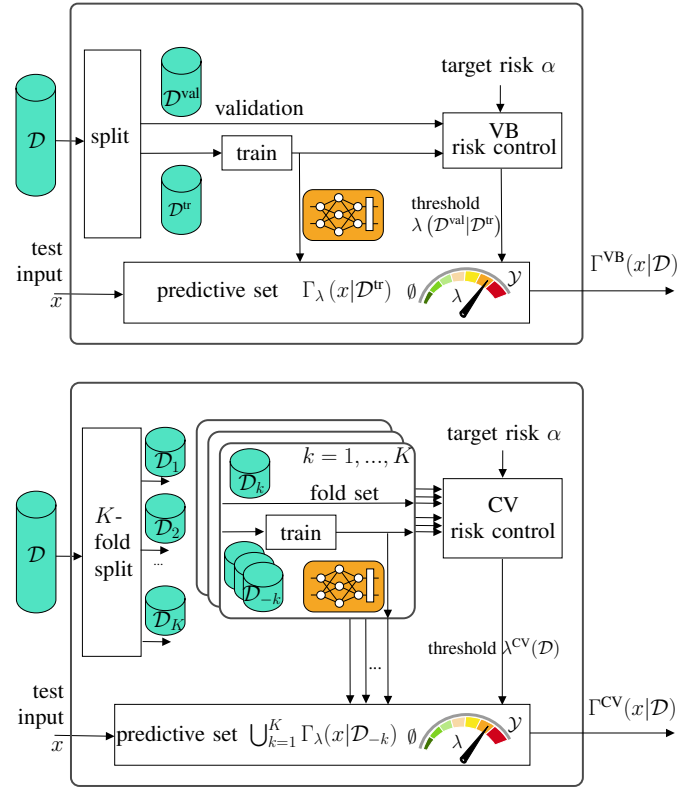

    \centering
    \includestandalone[trim=0.3cm 0.0cm 0.4cm 0cm, clip,width=8.9cm]{Figs/fig_tikz_crc_overview}
    \caption{ Illustration of {(top)} the existing validation-based conformal risk control (VB-CRC) \cite{angelopoulos2022conformal}; and {(bottom)} the proposed method cross-validation-based conformal risk control (CV-CRC), which aims at reducing the predictive sets sizes by reusing the available data $\D$ more efficiently.}
    \label{fig: fig_tikz_crc_overview}
\end{figure}

To elaborate, assume access to a data set $\mathcal{D}$ of $N$ pairs of examples consisting of input $x$ and output $y$. Based on the data set $\mathcal{D}$ and on a class of point predictors, CP and CRC produce a set predictor $\Gamma(x|\mathcal{D})$ mapping a test input $x$ into a subset of the output space. The size of the set predictor $\Gamma(x|\mathcal{D})$ provides a measure of the uncertainty of the predictor for input $x$ \cite{angelopoulos2023conformal}. On average over the data set $\D$ and over a test input-output pair $(x,y)$, we wish to guarantee the calibration condition 
\begin{equation}
    \E_{\rvD,\rv{x},\rv{y} \sim p_0(\D,x,y)} \big[ \ell\big(\rv{y},\Gamma(\rv{x}|\rvD) \big) \big] \leq \alpha, \label{eq: intro problem}
\end{equation}
where boldface fonts denote random quantities, $\ell(\cdot,\cdot)$ is a loss measure, and $\alpha$ a user-specified maximum average loss level. In (\ref{eq: intro problem}),  under the joint distribution $p_0(\D,x,y)$, the examples in the data set $\rvD$ and the test pair $(\rv{x},\rv{y})$ are assumed to be independent identically distributed (i.i.d.), or, more generally \emph{exchangeable}. 

CRC can satisfy the requirement \eqref{eq: intro problem} for any user-specified target average loss level $\alpha$, as long as the loss function is bounded and it decreases  as the predicted set grows. Examples of such loss functions are the 0-1 miscoverage probability 
\begin{equation}
    \ell(y,\Gamma)=\indicator(y \notin \Gamma), \label{eq:miscoverage}
\end{equation} 
which returns 1 if the true label $y$ is not in the set $\Gamma$ and 0 otherwise, and the \emph{false negative rate}, which returns the fraction of true values of $y$ that are not included in set $\Gamma$ for multi-label problems \cite{angelopoulos2022conformal}.

The requirement \eqref{eq: intro problem}  can be always satisfied for such monotonic loss functions by returning as set predictor $\Gamma$ the entire set of possible values for the output variable $y$. However,  a set predictor is useful only as long as it is of moderate average size. The motivation of this work is to construct a set predictor that meets \eqref{eq: intro problem}, while producing small predictive sets even in the presence of a limited data set $\mathcal{D}$. 

\subsection{State of the Art}

CP addresses the design of set predictors satisfying the calibration condition (\ref{eq: intro problem}) in the special case of the miscoverage loss (\ref{eq:miscoverage}) \cite{vovk2005algorithmic, fontana2022conformal, angelopoulos2023conformal}. There are several variants of CP, including validation-based CP (VB-CP), \emph{cross-validation-based CP (CV-CP)} \cite{barber2021predictive}, and full CP \cite{vovk2005algorithmic}. While full CP is considered to be impractical, requiring many rounds of retraining, VB-CP splits the data set into training and validation data sets, and it operates over a single round of training. However, the need to devote a separate data set for validation can significantly reduce the quality of the trained model, resulting in predictive sets of large sizes when data are limited \cite{barber2021predictive}.

CV-CP reduces the computational complexity as compared to full CP, while reducing the predicted set size as compared to VB-CP. This is done by partitioning the available data set into multiple folds, each acting as a validation data set for the model trained based on leave-fold-out data. At the cost of increasing the complexity, requiring as many training rounds as the number of folds, CV-CP was shown to produce important savings in terms of prediction set sizes \cite{cohen2022calibrating,deutschmann2023adaptive,gupta2022nested}.

Other extensions of CP include CP-aware training strategies \cite{stutz2021learning, einbinder2022training}, prediction under distributional shifts \cite{tibshirani2019conformal}, improvements in the training algorithms \cite{yang2021finite,kumar2018trainable}, novel calibration metrics \cite{holland2020making,perez2022beyond}, applications to engineering problems \cite{cohen2022calibrating,park2023quantum}, and online versions \cite{gibbs2021adaptive, feldman2022achieving} with applications \cite{zhang2023bayesianCP,cohen2023guaranteed}. 


CRC generalizes  CP to address the calibration criterion (\ref{eq: intro problem}) for a wider class of risks, with the only constraints that the risk function be bounded and monotonic in the predicted set size \cite{angelopoulos2022conformal,bates2021distribution,angelopoulos2021learn,feldman2022achieving}.  The original CRC is validation-based, and hence it may be referred to as VB-CRC for consistency with the terminology applied above for CP. Accordingly, it relies on a split of the data set into training  and validation sets, resulting in inefficient predictive sets when data are limited. 

\subsection{Main Contributions}

In this paper, we introduce a novel version of CRC based on cross-validation. The proposed CV-CRC method generalizes CV-CP, supporting arbitrary bounded and monotonic risk functions. As we will demonstrate, the design and analysis of CV-CRC are non-trivial extensions of CV-CP, requiring new definitions and proof techniques.  

The rest of the paper is organized as follows.  Sec.~\ref{sec: Conformal Risk Control Background} provides the necessary background, while CV-CRC is presented in Sec.~\ref{sec: Cross-Validation Conformal Risk Control}. Numerical experiments are reported in Sec.~\ref{sec: CV-CRC Experiment}, and Sec.~\ref{sec: Conclusion} draws some conclusions. All proofs are deferred to the supplementary material.

\section{Background}\label{sec: Conformal Risk Control Background}

Consider $N+1$ data points
\begin{equation}
    \underbrace{(\rv{x}[1],\rv{y}[1])}_{=\rv{z}[1]}\,\, ,\,\, \underbrace{(\rv{x}[2],\rv{y}[2])}_{=\rv{z}[2]} \,\, , \,\,\dots\,\, , \,\, \underbrace{\rv{x}[N+1],\rv{y}[N+1])}_{=\rv{z}[N+1]}
\end{equation} 
over the sample space $\mathcal{X} \times \mathcal{Y}$ that are drawn according to an \emph{exchangeable} joint distribution $p_0(\D,x,y)$ over index $i=1,\dots,N$. The first $N$ data points constitute the data set $\D=\{z[i]=(x[i],y[i])\}_{i=1}^N$, while the last data point $z[N+1]$ is the test pair, which is also denoted as $z=(x,y)$.  We fix a loss function $\ell:\mathcal{Y}\times 2^\mathcal{Y} \to \mathbb{R}$, which, given any label $y\in\mathcal{Y}$ and a predictive set $\Gamma\subseteq\mathcal{Y}$, returns a loss bounded as
\begin{equation}
    b \leq \ell(y,\Gamma)\leq B \label{eq: ell <= B}
\end{equation}
for some constants $B<\infty$ and $b\in\{-\infty\}\cup\mathbb{R}$. We further require that the loss is monotonic in the predictive set $\Gamma$ in the sense that the following implication holds
\begin{equation}
    \Gamma_1 \subseteq \Gamma_2 \quad\Rightarrow\quad \ell(y,\Gamma_1) \geq \ell(y,\Gamma_2) \quad \text{ for each } y\in\mathcal{Y}. \label{eq: nesting loss}
\end{equation}
Note that the 0-1 miscoverage loss (\ref{eq:miscoverage}) assumed by CP satisfies \eqref{eq: ell <= B} with $b=0$ and $B=1$, and it also satisfies the implication  \eqref{eq: nesting loss}.

For a given data set $\D$, VB-CRC uses a two-step procedure to satisfy the constraint \eqref{eq: intro problem} for some target average loss $\alpha$ in the interval \begin{equation} b~\leq~\alpha~\leq~B.\end{equation} To start, as illustrated in the top panel of Fig.~\ref{fig: fig_tikz_crc_overview}, the available data set $\D$ is split into $\Ntr$ examples forming the \emph{training set} $\Dtr$ and $\Nval=N-\Ntr$ points forming the \emph{validation set} $\Dval$ with $\D= \Dtr \cup \Dval$. In the first step of VB-CRC, a model is trained based on the training set $\Dtr$ using any arbitrary scheme. Then, in the second step, VB-CRC determines a threshold $\lambda\in\mathbb{R}$ by using the validation data set $\Dval$. As explained next, the threshold $\lambda$ dictates which labels $y\in\mathcal{Y}$ are to be included in the prediction set $\Gamma_\lambda(x|\Dtr)$ for any test input $x$ as follows.

A \emph{nonconformity (NC) score} $\NC((x,y)|\Dtr)$ is selected that evaluates the loss of the trained predictor on a pair $(x,y)$. Examples of NC scores include the residual between the label and a trained predictor for regression problems and the log-loss for classification problems \cite{angelopoulos2021uncertainty, romano2019conformalized, angelopoulos2023conformal}. With the given NC score, the set prediction is obtained as
\begin{equation}
     \Gamma_{\lambda} (x|\Dtr) = \Big\{ y^\prime\in\mathcal{Y} \Big|  \NC((x,y^\prime)|\Dtr) \leq \lambda \Big\}, \label{eq: def Gamma_lambda}
\end{equation}
thus including all labels $y^\prime\in\mathcal{Y}$ with NC score smaller or equal to the threshold $\lambda$. By design, the set \eqref{eq: def Gamma_lambda} satisfies the nesting property 
\begin{equation}
    \lambda_1<\lambda_2 \quad\Rightarrow\quad \Gamma_{\lambda_1} (x|\Dtr) \subseteq \Gamma_{\lambda_2}(x|\Dtr) \label{eq: nesting Gamma}
\end{equation}
for any input $x$ and data sets $\Dtr$. 

We define the \emph{risk} as the population, or test, loss of the predicted set \eqref{eq: def Gamma_lambda} as
\begin{equation}
    R(\lambda|\Dtr) = \E_{\rv{x},\rv{y}\sim p_0(x,y)} \Big[ \ell\big( \rv{y},\Gamma_\lambda(\rv{x}|\Dtr)\big) \Big]. \label{eq: ground-truth risk}
\end{equation}
Given the validation data set $\Dval=\{(\xval[i],\yval[i])\}_{i=1}^\Nval$, the risk \eqref{eq: ground-truth risk} can be estimated as
\begin{equation}
    \hat{R}^\text{val}(\lambda|\Dtr,\Dval) = \tfrac{1}{\Nval+1} \bigg( \sum_{i=1}^\Nval \ell\big(\yval[i], \Gamma_\lambda(\xval[i]|\Dtr) \big) + B \bigg), \label{eq: Rhat_val}
\end{equation}
which is a function of the threshold $\lambda$. This corresponds to a regularized, biased, empirical estimate of the risk \eqref{eq: ground-truth risk} that effectively adds an $(N+1)$-th dummy validation example with maximal loss $B$. 

VB-CRC chooses the lowest threshold $\lambda$ such that the  estimate \eqref{eq: Rhat_val} is no larger than the target average risk $\alpha$ as in 
\begin{equation}
     \lambda^\text{VB}(\Dval|\Dtr) = \inf_\lambda \Big\{\lambda \Big| \hat{R}^\text{val}(\lambda|\Dtr,\Dval) \leq \alpha  \Big\}. \label{eq: lambda_CRC}
\end{equation}
With this threshold choice, as proven in \cite{angelopoulos2022conformal}, the set predictor \eqref{eq: def Gamma_lambda} obtained via VB-CRC, i.e.,
\begin{equation}
    \Gamma^\text{VB}(x|\Dtr,\Dval) = \Gamma_{\lambda^\text{VB}(\Dval|\Dtr)}(x|\Dtr) \label{eq: VB-CRC Gamma}
\end{equation}
 ensures the desired condition (\ref{eq: intro problem}).  More precisely, the condition  (\ref{eq: intro problem}) holds for any fixed training set $\Dtr$, i.e., we have the inequality
\begin{equation}
    \E_{\rvDval,\rv{x},\rv{y} \sim p_0(\Dval,x,y)} \big[ \ell\big(\rv{y},\Gamma^\text{VB}(\rv{x}|\Dtr,\rvDval) \big) \big] \leq \alpha . \label{eq: barR(lambda|Dtr) <= alpha}
\end{equation}
Furthermore, in order for \eqref{eq: barR(lambda|Dtr) <= alpha} to hold, VB-CRC only requires the validation data $\Dval$ and test pair $(x,y)$ to be exchangeable.
 

\section{Cross-Validation Conformal Risk Control}\label{sec: Cross-Validation Conformal Risk Control}

While VB-CRC reviewed in the previous section guarantees the average risk condition \eqref{eq: barR(lambda|Dtr) <= alpha}, splitting the available data set into training and validation sets may potentially lead to inefficient set predictors, having large predictive sets on average. In this section, we introduce the proposed CV-CRC scheme that aims at improving the efficiency of VB-CRC \cite{angelopoulos2022conformal} via cross-validation \cite{barber2021predictive}, while still guaranteeing condition \eqref{eq: intro problem}.

To start, as illustrated in the bottom panel of Fig.~\ref{fig: fig_tikz_crc_overview}, the available data set $\D=\{z[i]\}_{i=1}^N$ is partitioned using a fixed mapping into $K$ \emph{folds} $\D = \{ \mathcal{D}_k\}_{k=1}^K$ of $N/K$-samples each, which is assumed to be an integer. We will write each $k$-th fold as $\mathcal{D}_k=\{(x_k[1],y_k[1]), \dots, (x_k[N/K],y_k[N/K])\}$, and we will denote the mapping of the $i$-th data point $z[i]$ to its fold index as $k[i]:\{1,\dots,N\}\to\{1,\dots,K\}$. Like VB-CRC, CV-CRC operates in two steps.

In the first step, for any $k$-th fold, a model is trained using the leave-fold-out training set $\D_{-k}=\D \setminus \mathcal{D}_k$ of $N-N/K$ samples. Accordingly, unlike VB-CRC, $K$ training rounds are required for CV-CRC. In the second step, as we will detail, CV-CRC determines a threshold $\lambda$ to determine which values of the output $y$ to include in the predicted set.

Given a threshold $\lambda$, CV-CRC produces the predictive set 
\begin{equation}
    \Gamma_{\lambda}^\text{CV} (x|\D) 
    = \Big\{ y^\prime\in\mathcal{Y} \Big|  \min_{k\in\{1,\dots,K\}} \big\{ \NC((x,y^\prime)|\D_{-k})  \big\}\leq \lambda \Big\}, \label{eq: def Gamma_lambda_CV}
\end{equation}
which includes all labels $y^\prime\in\mathcal{Y}$ with minimum, i.e., best case, NC score across the $K$ folds, that is not larger than $\lambda$.

To determine the threshold $\lambda$, CV-CRC estimates the population risk \eqref{eq: ground-truth risk} using cross-validation as
\begin{equation}
    \hat{R}^{\text{CV}}(\lambda|\D) 
     = \tfrac{1}{K+1} \bigg( \sum_{k=1}^K \!\tfrac{K}{N}\! \sum_{j=1}^{N/K} \ell\big(y_k[j], \Gamma_\lambda\big(x_k[j]\big|\D_{-k}\big) \big)  + B\bigg) . \label{eq: Rhat K-CV(lambda|D)}
\end{equation}
The cross-validation-based estimate \eqref{eq: Rhat K-CV(lambda|D)} can be interpreted as the conventional cross-validation loss evaluated on an augmented data set\begin{equation}
    \D^\text{aug} = \big\{ \underbrace{\D_1, \D_2 ,\dots, \D_K}_{=\D} , \D_\text{dummy} \big\}, \label{eq: augmented with dummy}
\end{equation} with the first $K$ folds being the available data set $\D=\{\D_1,\dots,\D_K\}$, and the additional ($K+1$)-th fold containing $N/K$ dummy points with the maximal loss of $B$. In a manner similar to VB-CRC, the addition to dummy data points acts as a regularizer for the estimate (\ref{eq: Rhat K-CV(lambda|D)}), which is required to provide performance guarantees. 

Finally, CV-CRC selects the threshold $\lambda$ by imposing that the cross-validation based estimate (\ref{eq: Rhat K-CV(lambda|D)}) of the loss is no larger than the target average loss value $\alpha$ as in 
\begin{equation}
    \lambda^\text{CV}(\D) 
     = \inf_\lambda \Big\{\lambda \Big| \hat{R}^{\text{CV}}(\lambda|\D) \leq \alpha  \Big\} . \label{eq: lambda K-CV-CRC(D)}
\end{equation}

CV-CRC reduces to the jackknife-minmax scheme in \cite{barber2021predictive} when evaluated with the miscoverage loss (\ref{eq:miscoverage}) in the special case of $K=N$ folds.

\begin{theorem}\label{thm: K-CV-CRC} 
   Fix any bounded and monotonic loss function $\ell(\cdot,\cdot)$ satisfying conditions \eqref{eq: ell <= B}  and  \eqref{eq: nesting loss}, and any NC score $\NC((x,y)|\Dtr)$ that is permutation-invariant with respect to the ordering of the examples in the training set $\Dtr$. For any number of folds satisfying $K \geq B/(\alpha-b)-1$,  the CV-CRC predictive set $\Gamma_{\lambda^\text{CV}(\D)}^\text{CV} (x|\D)$ with (\ref{eq: def Gamma_lambda_CV}) and \eqref{eq: lambda K-CV-CRC(D)} guarantees the condition
    \begin{equation}
        \E_{\rvD,\rv{x},\rv{y}\sim p_0(\D,x,y)} \Big[ \ell\big( \rv{y}, \Gamma^\text{CV}(\rv{x}|\rvD) \big) \Big] \leq \alpha. \label{eq: Rbar_K_CV def and <= alpha}
    \end{equation}
   
   
\end{theorem}
The theorem thus confirms that CV-CRC meets the desired condition (\ref{eq: intro problem}). In this regard, we note that, as in (\ref{eq: intro problem}), the average loss in \eqref{eq: Rbar_K_CV def and <= alpha} includes averaging over the entire data set $\D$, unlike the condition \eqref{eq: barR(lambda|Dtr) <= alpha} satisfied by VB-CRC. Furthermore, Theorem~\ref{thm: K-CV-CRC} requires the NC score to be permutation-invariant with respect to the data points in the training set, which is not the case for VB-CRC. Permutation-invariance is also needed for CV-CP \cite{barber2021predictive}, as well as for full CP \cite{vovk2005algorithmic}. In practice, a permutation-invariant NC score can be obtained by implementing permutation-invariant training schemes such as full gradient descent, in which the final trained model does not depend on the ordering of the training data points.

\section{Examples}\label{sec: CV-CRC Experiment}

In this section, we numerically validate the proposed CV-CRC using two synthetic examples. The first is a vector regression problem, whereas the second concerns the problem of temporal point process prediction \cite{omi2019fully,dubey2023bayesian}. Our code is publicly available\footnote{\url{https://github.com/kclip/cvcrc}}.

\subsection{Vector Regression}

Inspired by the example in \cite{barber2021predictive}, we first investigate a vector regression problem in which the output variable $y=[y_1,\dots,y_m]^\top$ is $m$-dimensional. The joint distribution of data set $\D$ and test pair $(x,y)$ is obtained as 
\begin{equation}
    p_0(\D,x,y) \!=\! \int p_0(\phi)  \bigg(\prod_{i=1}^{N+1}    p_0(x[i]) p_0(y[i]   |x   [i],\phi)  \bigg) \dd\phi  , \label{eq: p_0 (D Dte phi)}
\end{equation}
where $(x[N+1]=x,y[N+1]=y)$ is the test example, and we have the Gaussian distributions
\begin{subequations}
    \label{eq: Gaussian hierarchical model for CVCRC}
    \begin{eqnarray}
        p_0(x) &=& \Normdist(x|0,      d^{-1} I_d), \\
        p_0(y|x,\phi) &=& \Normdist(y|\phi^\top \cdot x, \beta_0^{-1} I_m), \label{eq: Gaussian hierarchical model p0(y|x,phi)}
    \end{eqnarray}
\end{subequations}
while $p_0(\phi)$ is a mixture of Gaussians with means determined by an i.i.d. Bernoulli vector $\rv{b}$ as
\begin{equation}
    p_0(\phi) = \E_{\rv{b}\stackrel{\text{i.i.d.}}{\sim}\Berndist(0.5)} \big[ \Normdist(\phi|\mu_0 \rv{b}, \gamma_0^{-1} I_d) \big].
\end{equation}
We set $\mu_0=10$, $\gamma_0=1$, $\beta_0=4$, $d=|\mathcal{X}|=50$, and $m=|\mathcal{Y}|=30$. Note that the distribution \eqref{eq: p_0 (D Dte phi)} is exchangeable.

\begin{figure}
    \centering
    \includegraphics[page=1,trim=3cm 9cm 4cm 10cm, clip,width= 8cm]{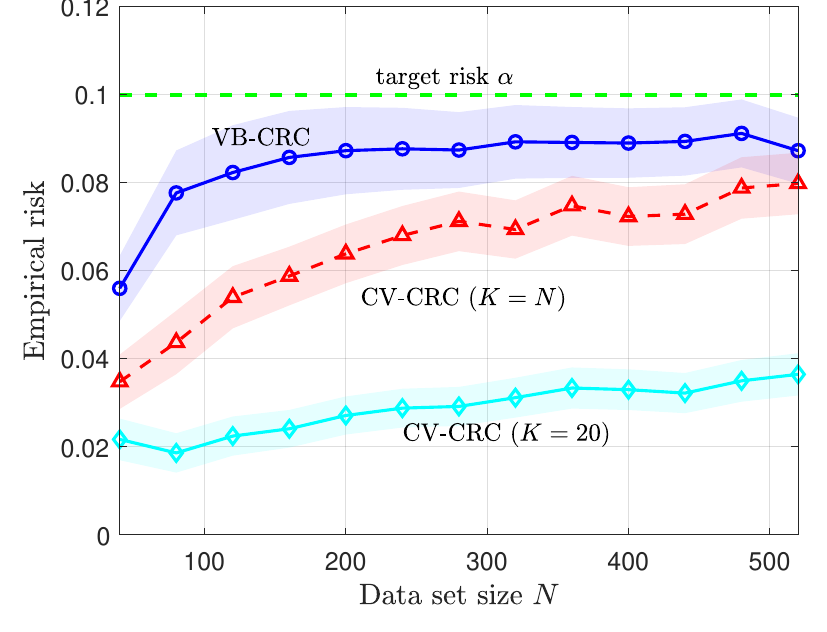}
    \caption{Empirical risk of VB-CRC and CV-CRC for the vector regression problem.}
    \label{fig: fig_toy_gaussian_risk_2023_10_03}
\end{figure}

\begin{figure}
    \centering
    \includegraphics[page=1,trim=3cm 9cm 4cm 10cm, clip,width= 8cm]{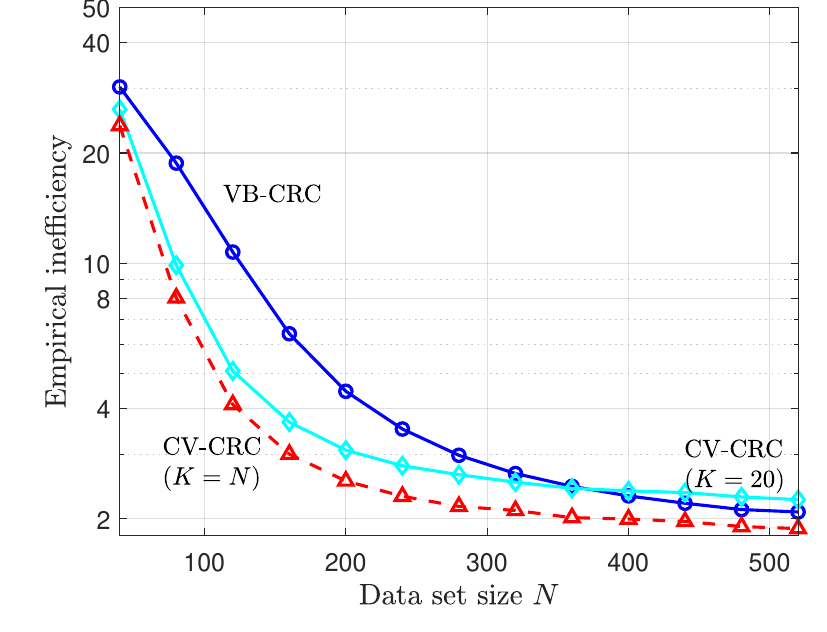}
    \caption{Empirical inefficiency of VB-CRC and CV-CRC  for the vector regression problem.}
    \label{fig: fig_toy_gaussian_inef_2023_10_03}
\end{figure}

Using maximum-likelihood learning, given a training data set $\Dtr$, we obtain the model parameter $\phi_\Dtr^\text{ML}$ used for the linear prediction model $\hat{y}(x|\Dtr)=(\phi_\D^\text{ML})^\top x$  as
$\phi_\Dtr^\text{ML} = X_\Dtr^\dagger Y_\Dtr$, where $(\cdot)^\dagger$ denotes the pseudo-inverse, $(\cdot)^\top$ denotes transpose, and the input and label data matrices $X_\D\in\mathbb{R}^{N\times d}$ and $Y_\D\in\mathbb{R}^{N\times m}$ have input $(x^\text{tr}[i])^\top$ and label $(y^\text{tr}[i])^\top$ as their $i$th rows, respectively.

The NC score is set to the maximum prediction residual across the $m$ dimensions of the output variable $y$ as
\begin{equation}
    \NC((x,y)|\Dtr)  = 2 \big\| y - \hat{y}(x|\Dtr)  \big\|_\infty,
\end{equation}
where the infinity norm $\big\|\cdot\big\|_{\infty}$ returns the largest magnitude of its input vector. This results in  predictive sets \eqref{eq: VB-CRC Gamma} and \eqref{eq: def Gamma_lambda_CV} with \eqref{eq: lambda K-CV-CRC(D)} in the form of $\Gamma=\Gamma_1\times\dots\times\Gamma_m$, with $\times$ being the Cartesian product and 
\begin{equation}
    \Gamma_j^\text{VB}=\Big\{y_j \, \Big| \, | y_j - [\hat{y}(x|\Dtr)]_j | \leq \lambda^\text{VB}(\Dval|\Dtr)/2 \Big\}
\end{equation}
with $[\cdot]_j$ standing for the $j$th element of its argument for VB-CRC, and
\begin{equation}
    \Gamma_j^\text{CV}= \bigcup_{k=1}^K \Big\{y_j \, \Big| \, | y_j - [\hat{y}(x|\D_{-k})]_j | \leq \lambda^\text{CV}(\D)/2 \Big\}  \label{eq: Gamma CV-CRC example 1}
\end{equation}
for CV-CRC. The loss function used in the risk \eqref{eq: intro problem} is defined as
\begin{equation}
    \ell(y,\Gamma) = \frac{1}{m} \sum_{j=1}^m \indicator\big(y_j \notin \Gamma_j\big), \label{eq: loss in risk example 1}
\end{equation}
which evaluates the fraction of entries of vector $y$ that are not included in the predictive set. This loss satisfies condition \eqref{eq: ell <= B} with $b=0$ and $B=1$. Note that CP is not applicable to this loss, since it is different from \eqref{eq:miscoverage}.

Lastly, we define the \emph{inefficiency} as the size of the predictive set evaluated as the average over all dimensions of the predictive intervals across the $m$ dimensions of the output $y$, i.e.,
$\mathrm{ineff}(\Gamma) = \tfrac{1}{m} \sum_{j=1}^m \big| \Gamma_j \big|$.

For target risk $\alpha=0.1$, the empirical risk and empirical inefficiency of $\Nte=200$ test covariate-output pairs, averaged over $50$ independent simulations, are shown in Fig.~\ref{fig: fig_toy_gaussian_risk_2023_10_03} and Fig.~\ref{fig: fig_toy_gaussian_inef_2023_10_03}. Fig.~\ref{fig: fig_toy_gaussian_risk_2023_10_03}, validates the theoretical result that CRC schemes satisfy condition \eqref{eq: intro problem}. However, from Fig.~\ref{fig: fig_toy_gaussian_inef_2023_10_03}, VB-CRC is observed to have  a larger inefficiency than CV-CRC, particularly in the small data set size regime. Thus, CV-CRC uses data more efficiently, with $K=20$ folds striking a good balance between inefficiency and computational complexity in this regime.

\subsection{Temporal Point Process Prediction}

\begin{figure}
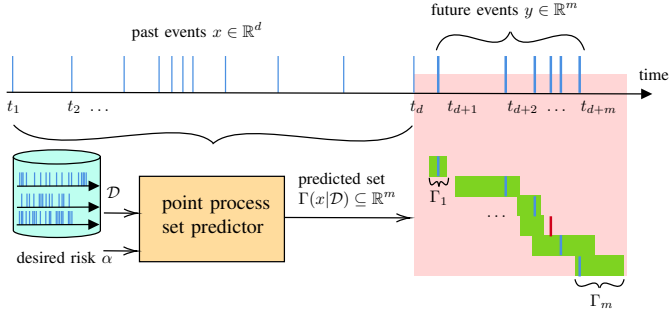

    \centering
    \includestandalone[trim=0.0cm 0.0cm 0.0cm 0.0cm, clip,width=9cm]{Figs/fig_tikz_point_process}
    \caption{Temporal point process prediction: After observing the past $d$ times $t_1,\dots,t_n$, a point process set predictor outputs predictive intervals $\Gamma_j(x|\D)$ for each of the next $m$ points with $j=1,\dots,m$.}
    \label{fig: fig_tikz_point_process}
\end{figure}

A temporal process consists of a sequence of events at random times $t_1,t_2,\ldots$ with $t_1<t_2<\ldots$ As illustrated in Fig.~\ref{fig: fig_tikz_point_process}, given the past $d$ events' timings $x=\{t_1,\dots,t_d\}$, the goal is to output intervals $\Gamma_j(x|\D)$ for each of the following $m$ events with $j=1,\ldots,m$. The loss function is defined as in \eqref{eq: loss in risk example 1}.


\begin{figure}
    \centering
    \includegraphics[page=1,trim=2.5cm 7.5cm 4cm 7cm, clip,width= 9cm]{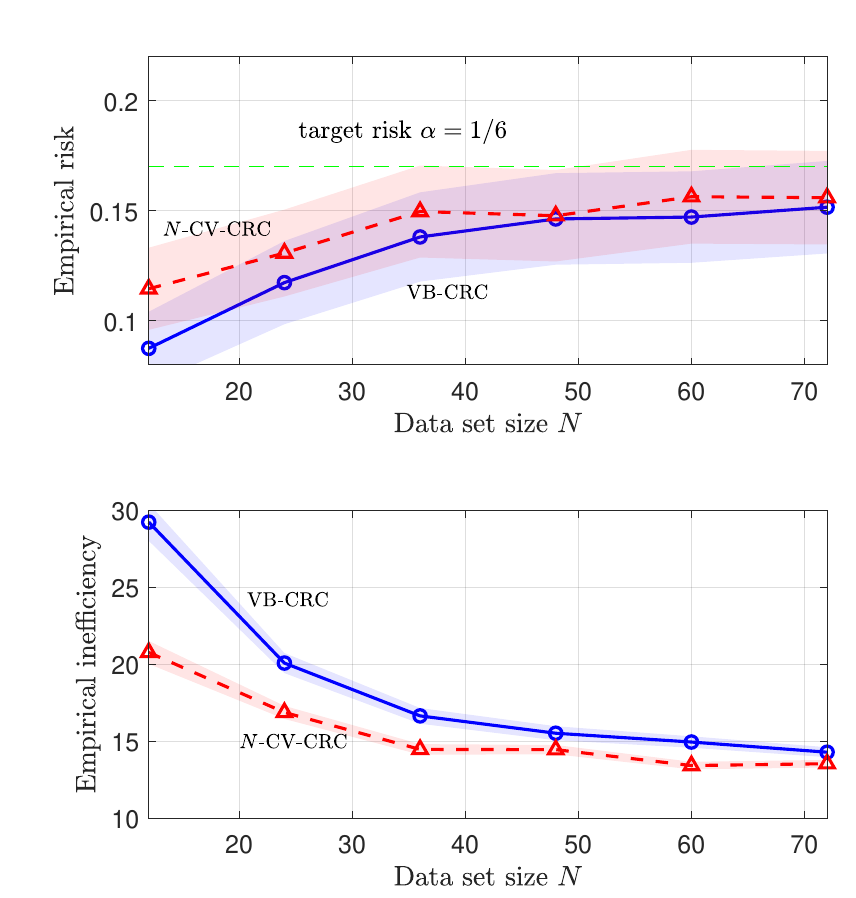}
    \caption{Empirical risk (top) and inefficiency (bottom) of VB-CRC and $N$-CV-CRC for the temporal point process prediction problem.}
    \label{fig: fig_hawkes2_risk_and_ineff_2024_01_2024}
\end{figure}

Data and test sequences of timings are generated following a self-exciting Hawkes process \cite{hawkes1971spectra} with intensity function
\begin{equation}
    \lambda(t|\mathcal{H}_t) = \mu + \sum_{i:t_i<t} \Big( \alpha_1 \beta_1 e^{-\beta_1(t-t_i)} + \alpha_2 \beta_2 e^{-\beta_2(t-t_i)} \Big), \nonumber
\end{equation}
with $\mu=0.2,\alpha_1=\alpha_2=0.4,\beta_1=1$ and $\beta_2=20$ \cite{omi2019fully}. The predictor is a recurrent neural network that outputs a predictive density function $p(t_{i+1}|t_1,\dots,t_i,\phi_{\Dtr})$ with trained parameter $\phi_\Dtr$ \cite{omi2019fully}.  The median $\hat{t}_{i+1}(t_1,\dots,t_i,\phi_\Dtr)$ of the predictive distribution is used as the point estimate for the $(i+1)$-th event. For $i>d$, estimates $\{\hat{t}_j\}_{j=d+1}^{i-1}$ are used in lieu of the correct timings in the point prediction.

VB-CRC \eqref{eq: VB-CRC Gamma} produces intervals
\begin{equation}
    \Gamma_j^\text{VB}=\Big\{y_j \, \Big| \, | y_j - \hat{t}_{d+j}(\Dtr) | \leq \gamma^j \lambda^\text{VB}(\Dval|\Dtr)/2 \Big\} ,
\end{equation}
where multiplication by the interval common ratio $\gamma=1.2$ increases the interval sizes for later predictions, and for the CV-CRC \eqref{eq: def Gamma_lambda_CV}, we have
\begin{equation}
    \Gamma_j^\text{CV}= \bigcup_{k=1}^K \Big\{y_j \, \Big| \, | y_j - \hat{t}_{d+j}(\D_{-k}) | \leq \gamma^j \lambda^\text{CV}(\D)/2 \Big\} .
\end{equation}
We set the length of the observed sequence as $d=60$, and predict the next $m=6$ events. We allow one event on average to lie outside the predicted intervals, i.e., $\alpha=1 / 6$. We average over $200$ independent simulations with $\Nte=1000$ test points in each run.

The top panel of Fig.~\ref{fig: fig_hawkes2_risk_and_ineff_2024_01_2024} illustrate the test risk \eqref{eq: loss in risk example 1} as function of data set size $N$, validating that both scheme attain risks lower than the desired level $\alpha$. The bottom panel of the figure shows that CV-CRC with $K=N$ reduces the average size of the predicted intervals.

\section{Conclusion}\label{sec: Conclusion}

In this paper, we have introduced a novel conformal risk control (CRC) scheme based on cross-validation, generalizing cross-validation CP to losses beyond miscoverage. The proposed CV-CRC was shown to provably control the average risk, with experiments demonstrating it to be more efficient than VB-CRC when the available data for training and calibration are scarce. Further work may consider using the jackknife+ of \cite{barber2021predictive} instead of the jackknife-minmax for more efficient predictive sets; and extending the scheme to meta-learning \cite{park2023fewshot}.



\clearpage 

\IEEEtriggeratref{18} 

\bibliographystyle{IEEEtran}
\bibliography{my_bib}

\ifcondition

\newpage
\onecolumn

\begin{center}
    \Huge{Cross-Validation Conformal Risk Control: Supplementary Material}
\end{center}

\appendices

\section{Proof that VB-CRC achieves target risk}\label{appendix: proof of lambda_CRC}

In this appendix, we prove condition \eqref{eq: barR(lambda|Dtr) <= alpha} for VB-CRC. While this result was originally shown in \cite{angelopoulos2022conformal}, here we provide an equivelant proof that is more convenient to support the  proof of Theorem~\ref{thm: K-CV-CRC} in  Appendix~\ref{appendix: proof of main theorem}. We start by bounding the VB-CRC threshold \eqref{eq: lambda_CRC} using the following steps
\begin{eqnarray}
     \lambda^\text{VB}(\Dval|\Dtr) 
     &=& \inf_\lambda \Bigg\{\lambda \Bigg| \tfrac{1}{\Nval+1} \bigg( \sum_{i=1}^\Nval \ell\big(\yval[i], \Gamma_\lambda(\xval[i]|\Dtr) \big) + B \bigg) \leq \alpha  \Bigg\} \nonumber \\
     &\geq& \inf_\lambda \Bigg\{\lambda \Bigg| \tfrac{1}{\Nval+1} \bigg( \sum_{i=1}^\Nval \ell\big(\yval[i], \Gamma_\lambda(\xval[i]|\Dtr) \big) + \ell\big(y, \Gamma_\lambda(x|\Dtr) \big) \bigg) \leq \alpha  \Bigg\}  \label{eq: lambda_prime} \\
     &=:& \lambda^\prime(\Dval,x,y|\Dtr) , \nonumber
\end{eqnarray}
where the inequality in \eqref{eq: lambda_prime} follows from \eqref{eq: ell <= B}. The ground-truth risk averaged over test example $(x,y)$ and validation set $\Dval$ is upper bounded as
\begin{subequations}
    \begin{eqnarray}
        \E_{\rvDval,\rv{x},\rv{y}\sim p_0(\Dval,x,y)} \Big[ \ell\big( \rv{y},\Gamma_{\lambda^\text{VB}(\rvDval                                |\Dtr)}(\rv{x}      |\Dtr)\big)\Big]   \label{eq: Rbar <= alpha (a)}
        &\quad\leq&            \E_{\rvDval,\rv{x},\rv{y}\sim p_0(\Dval,x,y)} \Big[ \ell\big( \rv{y},       \Gamma_{\lambda^\prime    (\rvDval,\rv{x},\rv{y}                  |\Dtr )}(\rv{x}     |\Dtr)\big)\Big]  \label{eq: Rbar <= alpha (b)} \\
        &\quad\leq& \alpha ,
    \end{eqnarray}
\end{subequations}
where the first inequality \eqref{eq: Rbar <= alpha (a)} follows the nesting property \eqref{eq: nesting Gamma} given inequality \eqref{eq: lambda_prime}. The second inequality \eqref{eq: Rbar <= alpha (b)} is an application of the following lemma, whose proof is deferred to Appendix~\ref{appendix: proof of lemma: conditional expectation is empirical mean}.
\begin{lemma}\label{lemma: conditional expectation is empirical mean}
    Let $\rv{v}_1,\dots,\rv{v}_M$ be random variables with an exchangeable joint distribution such that the equation $ \Prob \big( \tfrac{1}{M} \sum_{i=1}^M \rv{v}_i \leq \alpha \big) = 1$ holds. Then, we have the inequality $\E_{\rv{v}_{1:M}\sim p_0(v_{1:M})} [ \rv{v}_m] \leq \alpha$ for all $m \in \{1,...,M\}$.
\end{lemma}
To apply Lemma~\ref{lemma: conditional expectation is empirical mean} in \eqref{eq: Rbar <= alpha (b)}, we define $M=\Nval+1$ variables by
\begin{equation}
    \label{eq: v_i for lemma}
    \rv{v}_i = \begin{cases}
        \ell\big(\rv{y}^\text{val}[i], \Gamma_{\lambda^\prime(\rvDval,\rv{x},\rv{y}|\Dtr)}(\rv{x}^\text{val}[i]|\Dtr) \big)   & i=1,\dots,\Nval \\
        \ell\big(\rv{y}, \Gamma_{\lambda^\prime(\rvDval,\rv{x},\rv{y}|\Dtr)}(\rv{x}|\Dtr)                       & i=\Nval+1 , 
    \end{cases}
\end{equation}
whose empirical average is, by \eqref{eq: lambda_prime}, no greater than $\alpha$. Furthermore, to comply with the technical conditions of Lemma~\ref{lemma: conditional expectation is empirical mean}, variables $\rv{v}_{1:M}$ need to be exchangeable. This is justified by the following lemma, which is a corollary of \cite[Theorem~3]{kuchibhotla2020exchangeability} or \cite[Theorem~4]{dean1990linear}.
\begin{lemma}\label{lemma: dean1990linear}
    Let $\rv{w}_1,\dots,\rv{w}_M\in\mathcal{W}$ be a collection of exchangeable random vectors, $f:\mathcal{W}\to\mathbb{R}$ be a fixed mapping, and $g:\mathcal{W}^M\to\mathbb{R}$ be a fixed mapping that is permutation-invariant, i.e., oblivious to the ordering of its $M$ input values. Then, the $M$ random variables formed as $\rv{v}_1=f(\rv{w}_1,g(\rv{w}_{1:M})),\,\dots\,,\rv{v}_M=f(\rv{w}_M,g(\rv{w}_{1:M}))$ are exchangeable.
\end{lemma}
Lemma~\ref{lemma: dean1990linear} implies the exchangeability of variables \eqref{eq: v_i for lemma} by defining the $\Nval+1$ exchangeable vectors as
\begin{equation}
    \rv{w}_i = \begin{cases}
        \rv{z}^\text{val}[i] & i=1,\dots,\Nval \\
        (\rv{x},\rv{y})      & i=\Nval+1 ;
    \end{cases}
\end{equation}
the permutation invariant function is set as $g(\cdot)=\lambda^\prime(\cdot|\Dtr)$; the fixed mapping is
\begin{equation}
    \rv{v}_i=f\big(\rv{w}_i=(\rv{x}_i,\rv{y}_i),g(\rv{w}_{1:M})\big)=\ell(\rv{y}_i,\Gamma_{g(\rv{w}_{1:M})}(\rv{x}_i|\Dtr));
\end{equation}
and we focus on the average risk of the last term, i.e., $m=M=\Nval+1$. This completes the proof of \eqref{eq: barR(lambda|Dtr) <= alpha}.

\section{Proof of Lemma~\ref{lemma: conditional expectation is empirical mean} }\label{appendix: proof of lemma: conditional expectation is empirical mean}

In this appendix, we prove Lemma~\ref{lemma: conditional expectation is empirical mean}. To start, define a \emph{bag} $u= \Lbag u_1,\dots,u_M\Rbag$ of $M$ elements $u_1\dots,u_M$ as a multiset, i.e., as an unordered list with allowed repetitions \cite{vovk2005algorithmic}. By definition, two bags $u$ and $v$ are equal if they contain the same elements, irrespective of the ordering of their identical items, which we write as $u\stackrel{\text{bag}}{=}v$. One can form a bag out of a random vector $\rv{v}_1,\dots,\rv{v}_M \sim p_0(v_{1:M})$ by discarding the order of the items. Accordingly, the distribution of the bag $\rv{u}$ is given by
\begin{align}
    p_0(u) = \Prob\Big( \Lbag \rv{v}_1,...,\rv{v}_M \Rbag \stackrel{\text{bag}}{=} \Lbag u_1, \dots ,u_M \Rbag \Big) =  \sum_{\pi \in \Pi_M} \Prob\big(\rv{v}_1 = u_{\pi(1)},...,\rv{v}_M = u_{\pi(M)}\big), \label{eq: p_u of a bag}
\end{align}
where the sum is over the set $\Pi_M$ of all $M!$ permutations. For example, three Bernoulli variables $\rv{v}_1,\rv{v}_2,\rv{v}_3 \stackrel[\text{i.i.d.}]{}{\sim} \Berndist(q)$ with parameter $q\in[0,1]$ can constitute four different bags. In fact, bag $\rv{u} \stackrel{\text{bag}}{=} \Lbag \rv{v}_1,\rv{v}_2, \rv{v}_3 \Rbag $ equals $\Lbag 0,0 ,0 \Rbag$ with probability (w.p.) $(1-q)^3$, $\Lbag 0,0,1 \Rbag$ w.p. $3(1-q)^2 q$,  $\Lbag 0,1,1 \Rbag$ w.p. $3(1-q)q^2$, and  $\Lbag 1,1,1 \Rbag$ w.p. $q^3$.

With these definitions, we obtain the following chain of inequalities
\begin{subequations}
\label{eq: exchangeability lemma proof}
\begin{eqnarray}
    \E_{\rv{v}_{1:M}\sim p_0(v_{1:M})} [ \rv{v}_m]
    &=&  \E_{\rv{u} \sim p_0(u)} \Big[ \E_{\rv{v}_{1:M} \sim p_0(v_{1:M}|u)} \Big[ \rv{v}_m \Big| \Lbag \rv{v}_1,...,\rv{v}_M \Rbag \stackrel{\text{bag}}{=} \rv{u} \Big] \Big] \label{eq: exchangeability lemma proof (a)} \\
    &=&  \E_{\rv{u} \sim p_0(u)} \Big[ \tfrac{1}{M} \sum_{l=1}^M \rv{u}_l \Big] \label{eq: exchangeability lemma proof (b)} \\
    &=&  \E_{\rv{v}_{1:M} \sim p_0(v_{1:M})} \Big[ \E_{\rv{u} \sim p_0(u|v_{1:M})} \Big[ \tfrac{1}{M} \sum_{l=1}^M \rv{u}_l \Big| \rv{u} \stackrel{\text{bag}}{=} \Lbag \rv{v}_1, \dots ,\rv{v}_M \Rbag \Big] \Big] \label{eq: exchangeability lemma proof (c)} \\
    &=&  \E_{\rv{v}_{1:M} \sim p_0(v_{1:M})} \Big[ \E_{\rv{u} \sim p_0(u|v_{1:M})} \Big[ \tfrac{1}{M} \sum_{l=1}^M \rv{v}_l \Big| \rv{u} \stackrel{\text{bag}}{=} \Lbag \rv{v}_1, \dots ,\rv{v}_M \Rbag \Big] \Big] \label{eq: exchangeability lemma proof (d)} \\
    &=&  \E_{\rv{v}_{1:M} \sim p_0(v_{1:M})}  \Big[ \tfrac{1}{M} \sum_{l=1}^M \rv{v}_l \Big]  \label{eq: exchangeability lemma proof (e)} \\
    &\leq&  \alpha . \label{eq: exchangeability lemma proof (f)}
\end{eqnarray}
\end{subequations}
The inequalities of \eqref{eq: exchangeability lemma proof} are justified as follows: 
\eqref{eq: exchangeability lemma proof (a)} and \eqref{eq: exchangeability lemma proof (c)} stem from the law of iterated expectations over all possible bags of $M$ items;
\eqref{eq: exchangeability lemma proof (b)} arises from the fact that each item in the bag has an equal likelihood to be the realization of the $m$-th variable $\rv{v}_m$;
 is again the law of iterated expectation with the reintroduction of the random vector;
\eqref{eq: exchangeability lemma proof (d)} stems from the fact that if two bags have the same items,  their sum is identical;
\eqref{eq: exchangeability lemma proof (e)} leverages the fact that the bag given its random variables is a deterministically specified; 
and lastly, \eqref{eq: exchangeability lemma proof (f)} is by the assumption in Lemma~\ref{lemma: conditional expectation is empirical mean}. This concludes the proof of Lemma~\ref{lemma: conditional expectation is empirical mean}.

\section{Proof of Theorem~\ref{thm: K-CV-CRC}}\label{appendix: proof of main theorem}

To prove Theorem~\ref{thm: K-CV-CRC}, let us introduced an augmented data set, such that the last, $(K+1)$-th, fold $\Dte=\D_{K+1}$ is composed of $N/K$ arbitrary test points
\begin{equation}
    \tilde{\D} = \big\{ \underbrace{\D_1, \D_2 ,\dots, \D_K}_{=\D} , \underbrace{\D_{K+1}}_{=\Dte} \big\} \label{eq: augmented with test}
\end{equation}
with the test point $(x,y)$ included as the first point in the test set, i.e., $(x,y)=(\xte[1],\yte[1])=(x_{K+1}[1],y_{K+1}[1])$. By construction, all $N+N/K$ points in the augmented data set $\tilde{\D}$ are exchangeable and distributed according to joint distribution $p_0(\tilde{\D})=p_0(\D,\Dte)$. We denote the elements of the augmented set $\tilde{\D}$ in \eqref{eq: augmented with test} as
\begin{subequations}
    \begin{eqnarray}
        (\tilde{x}_k[j],\tilde{y}_k[j]) &=& (x_k[j],y_k[j])  \text{ for } k\in\{1,\dots K\} \\ 
        (\tilde{x}_{K+1}[j],\tilde{y}_{K+1}[j]) &=& (\xte[j],\yte[j]) . \label{eq: augmented data sets notations}
    \end{eqnarray}
\end{subequations}
Note that the augmented set $\tilde{\D}$ in \eqref{eq: augmented with test} is different than the augmented set using dummy points $\D^\text{aug}$ \eqref{eq: augmented with dummy}.
For a pair of folds indices $k^\prime,k\in\{1,\dots,K+1\}$ with $k\neq k^\prime$, we also define the augmented leave-two-folds-out (L2O) set as the augmented set without the two indexed folds, i.e.,
\begin{equation}
    \tilde{\D}_{-(k^\prime,k)}= \tilde{\D} \setminus \{ \mathcal{D}_{k^\prime} , \mathcal{D}_k \}. \label{eq: L2O}
\end{equation}
As a special case, when one of the indices points to the $(K+1)$-th fold, which is the test fold, the L2O reduces to the leave-one-out of the available data set $\tilde{\D}_{-(K+1,k)}=\D_{-k}$. For every fold within the augmented data set $\tilde{\D}$ \eqref{eq: augmented with test}, we evaluate the average L2O loss \eqref{eq: L2O}, minimized over the second fold index as
\begin{equation}
     \hat{R}_{\text{L2O}}^{\text{CV}}(\lambda|\tilde{\D}) = \tfrac{1}{K+1} \sum_{k=1}^{K+1} \tfrac{K}{N} \sum_{j=1}^{N/K} \min_{k^\prime\in\{1,\dots,K+1\}\setminus\{k\}} \Big\{ \ell\big(\tilde{y}_k[j], \Gamma_\lambda(\tilde{x}_k[j]|\tilde{\D}_{-(k,k^\prime)}) \big) \Big\} . \label{eq: Rhat_L2O def}
\end{equation}
Finally, we define the L2O threshold as the minimal threshold value for which the estimated average L2O risk \eqref{eq: Rhat_L2O def} is no larger than $\alpha$, i.e.,
\begin{equation}
     \lambda_{\text{L2O}}^{\text{CV}}(\tilde{\D}) = \inf_\lambda \Big\{\lambda \Big| \hat{R}_{\text{L2O}}^{\text{CV}}(\lambda|\tilde{\D}) \leq \alpha  \Big\} . \label{eq: lambda_primeKCV def}
\end{equation}
\begin{corollary}\label{corollary: lambda prime perm invariant}
    The L2O threshold $\lambda_{\text{L2O}}^{\text{CV}}(\tilde{\D})$ in \eqref{eq: lambda_primeKCV def} is fold-permutation-invariant, i.e., for any of the $(K+1)!$ possible fold-permutation mappings $\pi$, we have
    \begin{equation}
        \lambda_{\text{L2O}}^{\text{CV}}\big(\tilde{\D}\big) = \lambda_{\text{L2O}}^{\text{CV}}\big(\{\tilde{\D}_k\}_{k=1}^{K+1}\big) = \lambda_{\text{L2O}}^{\text{CV}}\big(\{\tilde{\D}_{\pi[k]}\}_{k=1}^{K+1}\big).
    \end{equation}
\end{corollary}
This is due to the commutative property of the outer fold-summation and of the inner, within-fold, summation in \eqref{eq: Rhat_L2O def}.
\begin{lemma}\label{lemma: lambdaL2O <= lambda} 
    The L2O threshold $\lambda_{\text{L2O}}^{\text{CV}}(\tilde{\D})$ in \eqref{eq: lambda_primeKCV def} lower bounds the $K$-CV-CRC threshold \eqref{eq: lambda K-CV-CRC(D)}
    \begin{equation}
        \lambda_{\text{L2O}}^{\text{CV}}(\tilde{\D}) \leq \lambda^{\text{CV}}(\D) .  \label{eq: lambda_CVCRC <= lambda_prime}
    \end{equation}
\end{lemma}
The proof of Lemma~\ref{lemma: lambdaL2O <= lambda}  is given in Appendix~\ref{appendix: proof of lemma: lambdaL2O <= lambda}. 

We now define $K+1$ random variables $\rv{v}_1,\dots,\rv{v}_{K+1}$, whose randomness stems from their dependence on the augmented data set $\tilde{\rvD}$. Each $k$-th random variable $\rv{v}_k$ is the minimal leave-two-fold-out empirical risk averaged over the $N/K$ examples in the validation fold $\rvD_k$, i.e.,
\begin{equation}
    \rv{v}_k = \tfrac{K}{N} \sum_{j=1}^{N/K} \min_{k^\prime\in\{1,\dots,K+1\}\setminus\{k\}} \Big\{ \ell\big( \tilde{\rv{y}}_k[j], \Gamma_{\lambda_{\text{L2O}}^{\text{CV}}(\tildervD)} (\tilde{\rv{x}}_k[j]   |\tildervD_{-(k^\prime,k)} )\big) \Big\} \quad\text{for}\quad k=1,\dots,K+1 . \label{eq: v_k mapping}
\end{equation}
The random variables $ \big\{ \rv{v}_1 , \dots , \rv{v}_{K+1} \big\} = \big\{ v_1(\lambda,\tildervD) , \dots , v_{K+1}(\lambda,\tildervD) \big\} $ are exchangeable for any fixed threshold due to the exchangeability of the folds in the augmented data set. Therefore, by Lemma~\ref{lemma: conditional expectation is empirical mean}, we have the inequality
\begin{equation}
    \E_{\tildervD\sim p_0(\tilde{\D})} \big[ v_{K+1} ( \lambda_{\text{L2O}}^{\text{CV}}(\tildervD), \tildervD)  \big] \leq \alpha . \label{eq: E_[v_{K+1}]<= alpha}
\end{equation}

We are now ready to follow the steps
\begin{subequations}
\begin{eqnarray}
    \E_{\rvD,\rv{x},\rv{y}\sim p_0(\D,x,y)} \Big[ \negspaceC& \ell\big(&\negspaceC \rv{y},       \Gamma^\text{CV}(\rv{x}|\rvD)    \big)\Big]     \nonumber \\
    \negspaceC&=&\negspaceC \E_{\rvD,\rv{x},\rv{y}\sim p_0(\D,x,y)} \Big[  \ell\big( \rv{y},       \bigcup_{k^\prime=1}^K\Gamma_{\lambda^\text{CV}       (\rvD                   )} (\rv{x}        |\rvD_{-k^\prime}       )\big)\Big] \label{eq: Rbar_K_CV <= alpha (a)} \\
    \negspaceC&\leq&\negspaceC \E_{\rvD,\rv{x},\rv{y}\sim p_0(\D,x,y)} \Big[  \min_{k^\prime\in\{1,\dots,K\}} \Big\{ \ell\big( \rv{y},       \Gamma_{\lambda^\text{CV}       (\rvD                   )} (\rv{x}        |\rvD_{-k^\prime}       )\big) \big\} \Big]  \label{eq: Rbar_K_CV <= alpha (b)} \\
    \negspaceC&=&\negspaceC \E_{\rvD,\rvDte\sim p_0(\D,\Dte)} \Big[  \min_{k^\prime\in\{1,\dots,K\}} \Big\{ \ell\big( \rvyte[1],       \Gamma_{\lambda^\text{CV}       (\rvD                   )} (\rvxte[1]        |\rvD_{-k^\prime}       )\big) \big\} \Big]  \label{eq: Rbar_K_CV <= alpha (c)} \\
    \negspaceC&=&\negspaceC \E_{\rvD,\rvDte\sim p_0(\D,\Dte)} \Big[  \tfrac{K}{N} \sum_{j=1}^{N/K} \min_{k^\prime\in\{1,\dots,K\}} \Big\{ \ell\big( \rvyte[j],       \Gamma_{\lambda^\text{CV}(\rvD)}         (\rvxte[j]         |\rvD_{-k^\prime}    )\big) \Big\} \label{eq: Rbar_K_CV <= alpha (d)} \\
    \negspaceC&=&\negspaceC \E_{\tildervD\sim p_0(\tilde{\D})} \Big[  \min_{k^\prime\in\{1,\dots,K\}} \Big\{ \tfrac{K}{N} \sum_{j=1}^{N/K} \ell\big( \tilde{\rv{y}}_{K+1}[j],       \Gamma_{\lambda^\text{CV}(\rvD)}         (\tilde{\rv{x}}_{K+1}[j]         |\tildervD_{-(k^\prime,K+1)}    )\big) \Big\} \Big]       
    \label{eq: Rbar_K_CV <= alpha (e)} \\
    \negspaceC&\leq&\negspaceC \E_{\tildervD\sim p_0(\tilde{\D})} \Big[  \min_{k^\prime\in\{1,\dots,K\}} \Big\{ \tfrac{K}{N} \sum_{j=1}^{N/K} \ell\big( \tilde{\rv{y}}_{K+1}[j],       \Gamma_{\lambda_{\text{L2O}}^{\text{CV}}(\tildervD)}         (\tilde{\rv{x}}_{K+1}[j]         |\tildervD_{-(k^\prime,K+1)}    )\big) \Big\} \Big]       \label{eq: Rbar_K_CV <= alpha (f)} \\
    \negspaceC&\leq&\negspaceC \alpha, 
    \label{eq: Rbar_K_CV <= alpha (g)}
\end{eqnarray}
\end{subequations}
where \eqref{eq: Rbar_K_CV <= alpha (a)} is a consequence of \eqref{eq: def Gamma_lambda_CV}, which is equivalent to $\Gamma_{\lambda}^\text{CV} (x|\D) = \bigcup_{k=1}^K\Gamma_{\lambda} (x|\D_{-k})$; 
inequality \eqref{eq: Rbar_K_CV <= alpha (b)} is due to the nesting property \eqref{eq: nesting loss} applied on a particular left-fold-out $k^\prime$ which is a subset of the union of all left-fold-out sets; 
\eqref{eq: Rbar_K_CV <= alpha (d)} leverages exchangeability as all test points have the same expected loss; 
\eqref{eq: Rbar_K_CV <= alpha (e)} uses the augmented data set notations \eqref{eq: augmented data sets notations}; 
inequality \eqref{eq: Rbar_K_CV <= alpha (f)} is an outcome of the nesting properties \eqref{eq: nesting Gamma} and \eqref{eq: nesting loss} with inequality \eqref{eq: lambda_CVCRC <= lambda_prime}; 
in inequality \eqref{eq: Rbar_K_CV <= alpha (g)}, we have used \eqref{eq: E_[v_{K+1}]<= alpha}, alongside Corollary~\ref{corollary: lambda prime perm invariant}, stating that the L2O threshold is fold-invariant.
This completes the proof of Theorem~\ref{thm: K-CV-CRC}.

\section{Proof of Lemma~\ref{lemma: lambdaL2O <= lambda} }\label{appendix: proof of lemma: lambdaL2O <= lambda}
The proof of Lemma~\ref{lemma: lambdaL2O <= lambda} stated in Appendix~\ref{appendix: proof of main theorem} follows the steps
\begin{subequations}
\begin{eqnarray}
    \lambda_{\text{L2O}}^{\text{CV}}(\tilde{\D}) 
    &\negspaceC=\negspaceC& \inf_\lambda \Bigg\{\lambda \Bigg| \tfrac{1}{K+1} \sum_{k=1}^{K+1} \tfrac{K}{N} \sum_{j=1}^{N/K} \min_{k^\prime\in\{1,\dots,K+1\}\setminus\{k\}} \Big\{ \ell\big(\tilde{y}_k[j], \Gamma_\lambda(\tilde{x}_k[j]|\tilde{\D}_{-(k,k^\prime)}) \big) \Big\} \leq \alpha  \Bigg\} \label{eq: lambda_prime <= lambda proof (a)} \\
    &\negspaceC=\negspaceC& \inf_\lambda \Bigg\{\lambda \Bigg| \tfrac{1}{K+1} \Bigg( \sum_{k=1}^K \tfrac{K}{N} \sum_{j=1}^{N/K}  \min_{k^\prime\in\{1,\dots,K+1\}\setminus\{k\}} \Big\{  \ell\big(\tilde{y}_k[j], \Gamma_\lambda(\tilde{x}_k[j]|\tilde{\D}_{-(k,k^\prime)}) \big) \Big\} \label{eq: lambda_prime <= lambda proof (b)} \\
    &\negspaceC\negspaceC& \quad\quad\quad\quad\quad\quad\quad\quad+ \tfrac{K}{N} \sum_{j=1}^{N/K} \min_{k^\prime\in\{1,\dots,K\}} \Big\{ \ell\big(\tilde{y}_{K+1}[j], \Gamma_\lambda(\tilde{x}_{K+1}[j]|\tilde{\D}_{-(K+1,k^\prime)}) \big) \Big\} \Bigg) \leq \alpha  \Bigg\} \nonumber \\
    &\negspaceC\leq\negspaceC& \inf_\lambda \Bigg\{\lambda \Bigg| \tfrac{1}{K+1} \Big( \sum_{k=1}^K \tfrac{K}{N} \sum_{j=1}^{N/K} \min_{k^\prime\in\{1,\dots,K+1\}\setminus\{k\}} \Big\{ \ell\big(\tilde{y}_k[j], \Gamma_\lambda(\tilde{x}_k[j]|\tilde{\D}_{-(k,k^\prime)}) \big) \Big\} + \tfrac{K}{N} \sum_{j=1}^{N/K} B \Big) \leq \alpha  \Bigg\} \label{eq: lambda_prime <= lambda proof (c)} \\
    &\negspaceC\leq\negspaceC& \inf_\lambda \Bigg\{\lambda \Bigg| \tfrac{1}{K+1}  \Big( \sum_{k=1}^K \tfrac{K}{N} \sum_{j=1}^{N/K} \ell\big(\tilde{y}_k[j], \Gamma_\lambda(\tilde{x}_k[j]|\tilde{\D}_{-(k,K+1)}) \big) + B \Big) \leq \alpha  \Bigg\} \label{eq: lambda_prime <= lambda proof (d)} \\
    &\negspaceC=\negspaceC& \inf_\lambda \Bigg\{\lambda \Bigg| \tfrac{1}{K+1}  \Big( \sum_{k=1}^K  \tfrac{K}{N} \sum_{j=1}^{N/K} \ell\big(y_k[j], \Gamma_\lambda(x_k[j]|\D_{-k}) \big) + B \Big) \leq \alpha  \Bigg\} \label{eq: lambda_prime <= lambda proof (e)} \\
    &\negspaceC=\negspaceC& \lambda^\text{CV}(\D), \label{eq: lambda_prime <= lambda proof (f)}
\end{eqnarray}
\end{subequations}
where 
\eqref{eq: lambda_prime <= lambda proof (a)} stems from the definition in \eqref{eq: lambda_primeKCV def};
\eqref{eq: lambda_prime <= lambda proof (b)} is obtained by decomposing the first sum into its first $K$ summation terms, and by listing the last term, the $(K+1)$-th, on its own;
and \eqref{eq: lambda_prime <= lambda proof (f)} follows the definition in \eqref{eq: lambda K-CV-CRC(D)}.
This completes the proof of Lemma~\ref{lemma: lambdaL2O <= lambda}.

\fi

\end{document}

%% file: Figs/fig_tikz_crc_overview.tex

\tikzset{every picture/.style={line width=0.75pt}} 

\begin{tikzpicture}[x=0.75pt,y=0.75pt,yscale=-1,xscale=1]

\draw  [fill={rgb, 255:red, 80; green, 227; blue, 194 }  ,fill opacity=1 ] (198.45,189.78) -- (198.45,232.58) .. controls (198.45,240.09) and (189.49,246.18) .. (178.45,246.18) .. controls (167.4,246.18) and (158.45,240.09) .. (158.45,232.58) -- (158.45,189.78) .. controls (158.45,182.27) and (167.4,176.18) .. (178.45,176.18) .. controls (189.49,176.18) and (198.45,182.27) .. (198.45,189.78) .. controls (198.45,197.29) and (189.49,203.38) .. (178.45,203.38) .. controls (167.4,203.38) and (158.45,197.29) .. (158.45,189.78) ;
\draw  [fill={rgb, 255:red, 80; green, 227; blue, 194 }  ,fill opacity=1 ] (198.45,70.27) -- (198.45,113.07) .. controls (198.45,120.58) and (189.49,126.67) .. (178.45,126.67) .. controls (167.4,126.67) and (158.45,120.58) .. (158.45,113.07) -- (158.45,70.27) .. controls (158.45,62.76) and (167.4,56.67) .. (178.45,56.67) .. controls (189.49,56.67) and (198.45,62.76) .. (198.45,70.27) .. controls (198.45,77.78) and (189.49,83.87) .. (178.45,83.87) .. controls (167.4,83.87) and (158.45,77.78) .. (158.45,70.27) ;
\draw   (80.33,59.33) -- (137.33,59.33) -- (137.33,249.67) -- (80.33,249.67) -- cycle ;
\draw   (120.33,274.33) -- (531,274.33) -- (531,336) -- (120.33,336) -- cycle ;
\draw    (49.67,306.33) -- (116.67,306.97) ;
\draw [shift={(119.67,307)}, rotate = 180.55] [fill={rgb, 255:red, 0; green, 0; blue, 0 }  ][line width=0.08]  [draw opacity=0] (8.93,-4.29) -- (0,0) -- (8.93,4.29) -- cycle    ;
\draw    (298,163.67) -- (298.65,271.33) ;
\draw [shift={(298.67,274.33)}, rotate = 269.65] [fill={rgb, 255:red, 0; green, 0; blue, 0 }  ][line width=0.08]  [draw opacity=0] (8.93,-4.29) -- (0,0) -- (8.93,4.29) -- cycle    ;
\draw    (47.17,151.17) -- (77.33,151.62) ;
\draw [shift={(80.33,151.67)}, rotate = 180.86] [fill={rgb, 255:red, 0; green, 0; blue, 0 }  ][line width=0.08]  [draw opacity=0] (8.93,-4.29) -- (0,0) -- (8.93,4.29) -- cycle    ;
\draw  [fill={rgb, 255:red, 80; green, 227; blue, 194 }  ,fill opacity=1 ] (51.1,97.58) -- (51.1,193.71) .. controls (51.1,201.22) and (42.14,207.31) .. (31.1,207.31) .. controls (20.05,207.31) and (11.1,201.22) .. (11.1,193.71) -- (11.1,97.58) .. controls (11.1,90.07) and (20.05,83.98) .. (31.1,83.98) .. controls (42.14,83.98) and (51.1,90.07) .. (51.1,97.58) .. controls (51.1,105.09) and (42.14,111.18) .. (31.1,111.18) .. controls (20.05,111.18) and (11.1,105.09) .. (11.1,97.58) ;
\draw    (137,134) -- (441,134) ;
\draw [shift={(444,134)}, rotate = 180] [fill={rgb, 255:red, 0; green, 0; blue, 0 }  ][line width=0.08]  [draw opacity=0] (8.93,-4.29) -- (0,0) -- (8.93,4.29) -- cycle    ;
\draw   (219.67,146.24) -- (290.33,146.24) -- (290.33,180.88) -- (219.67,180.88) -- cycle ;

\draw    (290.33,163.67) -- (441,163.99) ;
\draw [shift={(444,164)}, rotate = 180.12] [fill={rgb, 255:red, 0; green, 0; blue, 0 }  ][line width=0.08]  [draw opacity=0] (8.93,-4.29) -- (0,0) -- (8.93,4.29) -- cycle    ;
\draw  [color={rgb, 255:red, 74; green, 74; blue, 74 }  ,draw opacity=1 ][line width=1.5]  (66.33,19.67) .. controls (66.33,15.62) and (69.62,12.33) .. (73.67,12.33) -- (577.26,12.33) .. controls (581.31,12.33) and (584.59,15.62) .. (584.59,19.67) -- (584.59,349.68) .. controls (584.59,353.73) and (581.31,357.01) .. (577.26,357.01) -- (73.67,357.01) .. controls (69.62,357.01) and (66.33,353.73) .. (66.33,349.68) -- cycle ;
\draw  [fill={rgb, 255:red, 245; green, 166; blue, 35 }  ,fill opacity=1 ] (301.63,211.1) .. controls (301.63,205.41) and (306.24,200.8) .. (311.93,200.8) -- (372.69,200.8) .. controls (378.39,200.8) and (383,205.41) .. (383,211.1) -- (383,242.03) .. controls (383,247.72) and (378.39,252.33) .. (372.69,252.33) -- (311.93,252.33) .. controls (306.24,252.33) and (301.63,247.72) .. (301.63,242.03) -- cycle ;
\draw [fill={rgb, 255:red, 255; green, 255; blue, 255 }  ,fill opacity=1 ]   (308.86,218.08) -- (318.96,218.08) ;
\draw [fill={rgb, 255:red, 255; green, 255; blue, 255 }  ,fill opacity=1 ]   (308.99,235.17) -- (319.1,235.17) ;
\draw [fill={rgb, 255:red, 255; green, 255; blue, 255 }  ,fill opacity=1 ]   (337.81,208.99) -- (357.18,218.05) ;
\draw [fill={rgb, 255:red, 255; green, 255; blue, 255 }  ,fill opacity=1 ]   (337.81,208.99) -- (357.18,235.3) ;
\draw [fill={rgb, 255:red, 255; green, 255; blue, 255 }  ,fill opacity=1 ]   (337.81,226.11) -- (357.18,235.3) ;
\draw [fill={rgb, 255:red, 255; green, 255; blue, 255 }  ,fill opacity=1 ]   (337.81,243.48) -- (357.18,218.05) ;
\draw [fill={rgb, 255:red, 255; green, 255; blue, 255 }  ,fill opacity=1 ]   (337.81,226.23) -- (357.18,218.05) ;
\draw [fill={rgb, 255:red, 255; green, 255; blue, 255 }  ,fill opacity=1 ]   (337.81,243.48) -- (357.18,235.3) ;
\draw  [fill={rgb, 255:red, 255; green, 255; blue, 255 }  ,fill opacity=1 ] (352.87,218.05) .. controls (352.87,215.75) and (354.8,213.89) .. (357.18,213.89) .. controls (359.55,213.89) and (361.48,215.75) .. (361.48,218.05) .. controls (361.48,220.35) and (359.55,222.22) .. (357.18,222.22) .. controls (354.8,222.22) and (352.87,220.35) .. (352.87,218.05) -- cycle ;
\draw  [fill={rgb, 255:red, 255; green, 255; blue, 255 }  ,fill opacity=1 ] (352.87,235.3) .. controls (352.87,233) and (354.8,231.13) .. (357.18,231.13) .. controls (359.55,231.13) and (361.48,233) .. (361.48,235.3) .. controls (361.48,237.6) and (359.55,239.46) .. (357.18,239.46) .. controls (354.8,239.46) and (352.87,237.6) .. (352.87,235.3) -- cycle ;
\draw [fill={rgb, 255:red, 255; green, 255; blue, 255 }  ,fill opacity=1 ]   (361.48,218.05) -- (377.68,218.05) ;
\draw [fill={rgb, 255:red, 255; green, 255; blue, 255 }  ,fill opacity=1 ]   (361.48,235.3) -- (377.68,235.3) ;
\draw  [fill={rgb, 255:red, 255; green, 255; blue, 255 }  ,fill opacity=1 ] (365.42,208.74) -- (372.56,208.74) -- (372.56,242.24) -- (365.42,242.24) -- cycle ;
\draw [fill={rgb, 255:red, 255; green, 255; blue, 255 }  ,fill opacity=1 ]   (337.84,209.15) -- (318.47,218.21) ;
\draw [fill={rgb, 255:red, 255; green, 255; blue, 255 }  ,fill opacity=1 ]   (337.84,209.15) -- (318.47,235.45) ;
\draw [fill={rgb, 255:red, 255; green, 255; blue, 255 }  ,fill opacity=1 ]   (337.84,226.26) -- (318.47,235.45) ;
\draw [fill={rgb, 255:red, 255; green, 255; blue, 255 }  ,fill opacity=1 ]   (337.84,243.63) -- (318.47,218.21) ;
\draw [fill={rgb, 255:red, 255; green, 255; blue, 255 }  ,fill opacity=1 ]   (337.84,226.39) -- (318.47,218.21) ;
\draw [fill={rgb, 255:red, 255; green, 255; blue, 255 }  ,fill opacity=1 ]   (337.84,243.63) -- (318.47,235.45) ;
\draw  [fill={rgb, 255:red, 255; green, 255; blue, 255 }  ,fill opacity=1 ] (314.89,218.05) .. controls (314.89,215.75) and (316.82,213.89) .. (319.2,213.89) .. controls (321.57,213.89) and (323.5,215.75) .. (323.5,218.05) .. controls (323.5,220.35) and (321.57,222.22) .. (319.2,222.22) .. controls (316.82,222.22) and (314.89,220.35) .. (314.89,218.05) -- cycle ;
\draw  [fill={rgb, 255:red, 255; green, 255; blue, 255 }  ,fill opacity=1 ] (314.89,235.3) .. controls (314.89,233) and (316.82,231.13) .. (319.2,231.13) .. controls (321.57,231.13) and (323.5,233) .. (323.5,235.3) .. controls (323.5,237.6) and (321.57,239.46) .. (319.2,239.46) .. controls (316.82,239.46) and (314.89,237.6) .. (314.89,235.3) -- cycle ;
\draw  [fill={rgb, 255:red, 255; green, 255; blue, 255 }  ,fill opacity=1 ] (333.5,208.99) .. controls (333.5,206.69) and (335.43,204.83) .. (337.81,204.83) .. controls (340.19,204.83) and (342.12,206.69) .. (342.12,208.99) .. controls (342.12,211.29) and (340.19,213.15) .. (337.81,213.15) .. controls (335.43,213.15) and (333.5,211.29) .. (333.5,208.99) -- cycle ;
\draw  [fill={rgb, 255:red, 255; green, 255; blue, 255 }  ,fill opacity=1 ] (333.5,226.23) .. controls (333.5,223.94) and (335.43,222.07) .. (337.81,222.07) .. controls (340.19,222.07) and (342.12,223.94) .. (342.12,226.23) .. controls (342.12,228.53) and (340.19,230.4) .. (337.81,230.4) .. controls (335.43,230.4) and (333.5,228.53) .. (333.5,226.23) -- cycle ;
\draw  [fill={rgb, 255:red, 255; green, 255; blue, 255 }  ,fill opacity=1 ] (333.5,243.48) .. controls (333.5,241.18) and (335.43,239.31) .. (337.81,239.31) .. controls (340.19,239.31) and (342.12,241.18) .. (342.12,243.48) .. controls (342.12,245.78) and (340.19,247.64) .. (337.81,247.64) .. controls (335.43,247.64) and (333.5,245.78) .. (333.5,243.48) -- cycle ;

\draw    (502.33,179.67) -- (502.98,271.67) ;
\draw [shift={(503,274.67)}, rotate = 269.6] [fill={rgb, 255:red, 0; green, 0; blue, 0 }  ][line width=0.08]  [draw opacity=0] (8.93,-4.29) -- (0,0) -- (8.93,4.29) -- cycle    ;
\draw    (501.46,95) -- (501.4,119.8) ;
\draw [shift={(501.4,122.8)}, rotate = 270.14] [fill={rgb, 255:red, 0; green, 0; blue, 0 }  ][line width=0.08]  [draw opacity=0] (8.93,-4.29) -- (0,0) -- (8.93,4.29) -- cycle    ;
\draw    (531,303.33) -- (674.67,303.33) ;
\draw [shift={(677.67,303.33)}, rotate = 180] [fill={rgb, 255:red, 0; green, 0; blue, 0 }  ][line width=0.08]  [draw opacity=0] (8.93,-4.29) -- (0,0) -- (8.93,4.29) -- cycle    ;
\draw    (137,164) -- (216.67,163.68) ;
\draw [shift={(219.67,163.67)}, rotate = 179.77] [fill={rgb, 255:red, 0; green, 0; blue, 0 }  ][line width=0.08]  [draw opacity=0] (8.93,-4.29) -- (0,0) -- (8.93,4.29) -- cycle    ;
\draw  [color={rgb, 255:red, 255; green, 255; blue, 255 }  ,draw opacity=1 ][fill={rgb, 255:red, 126; green, 211; blue, 33 }  ,fill opacity=1 ] (444.04,298.22) .. controls (440.28,300.35) and (436.89,303.06) .. (434,306.21) -- (429.13,303.08) .. controls (431.95,298.89) and (435.42,295.17) .. (439.41,292.04) -- (444.04,298.22) -- cycle ;
\draw  [color={rgb, 255:red, 255; green, 255; blue, 255 }  ,draw opacity=1 ][fill={rgb, 255:red, 184; green, 233; blue, 134 }  ,fill opacity=1 ] (453.59,294.65) .. controls (450,295.43) and (446.6,296.71) .. (443.48,298.43) -- (439.1,292.48) .. controls (442.67,289.65) and (446.7,287.37) .. (451.07,285.76) -- (453.59,294.65) -- cycle ;
\draw  [color={rgb, 255:red, 255; green, 255; blue, 255 }  ,draw opacity=1 ][fill={rgb, 255:red, 248; green, 217; blue, 166 }  ,fill opacity=1 ] (464.85,293.8) .. controls (464.03,293.75) and (463.21,293.73) .. (462.38,293.73) .. controls (459.36,293.73) and (456.42,294.05) .. (453.59,294.65) -- (451.1,286.03) .. controls (455.72,284.43) and (460.68,283.5) .. (465.85,283.36) -- (464.85,293.8) -- cycle ;
\draw  [color={rgb, 255:red, 255; green, 255; blue, 255 }  ,draw opacity=1 ][fill={rgb, 255:red, 248; green, 231; blue, 28 }  ,fill opacity=1 ] (466.79,283.08) .. controls (471.98,283.08) and (476.97,283.94) .. (481.63,285.51) -- (476.48,296.11) .. controls (472.87,294.74) and (469.02,293.88) .. (465,293.62) -- (466.02,283.09) .. controls (466.28,283.08) and (466.53,283.08) .. (466.79,283.08) -- cycle ;
\draw  [color={rgb, 255:red, 255; green, 255; blue, 255 }  ,draw opacity=1 ][fill={rgb, 255:red, 245; green, 166; blue, 35 }  ,fill opacity=1 ] (486.14,301.15) .. controls (483.14,298.99) and (479.8,297.22) .. (476.2,295.93) -- (481.63,285.51) .. controls (486.79,287.14) and (491.54,289.58) .. (495.69,292.67) -- (486.14,301.15) -- cycle ;
\draw  [color={rgb, 255:red, 255; green, 255; blue, 255 }  ,draw opacity=1 ][fill={rgb, 255:red, 208; green, 2; blue, 27 }  ,fill opacity=1 ] (495.78,311.76) .. controls (493.15,307.75) and (489.82,304.16) .. (485.93,301.15) -- (495.54,292.87) .. controls (500.79,296.73) and (505.23,301.46) .. (508.59,306.8) -- (495.78,311.76) -- cycle ;
\draw  [color={rgb, 255:red, 255; green, 255; blue, 255 }  ,draw opacity=1 ][fill={rgb, 255:red, 65; green, 117; blue, 5 }  ,fill opacity=1 ][line width=0.75]  (434.1,306.33) .. controls (431.2,309.5) and (428.82,313.14) .. (427.08,317.12) -- (423.14,316.22) .. controls (424.45,311.5) and (426.49,307.07) .. (429.13,303.08) -- (434.1,306.33) -- cycle ;
\draw  [draw opacity=0][line width=2.25]  (419.77,315.41) .. controls (424.59,296.17) and (443.19,281.55) .. (465.78,280.96) .. controls (485.47,280.45) and (502.7,290.77) .. (510.63,305.99) -- (466.92,325.07) -- cycle ; \draw  [color={rgb, 255:red, 155; green, 155; blue, 155 }  ,draw opacity=1 ][line width=2.25]  (419.77,315.41) .. controls (424.59,296.17) and (443.19,281.55) .. (465.78,280.96) .. controls (485.47,280.45) and (502.7,290.77) .. (510.63,305.99) ;  
\draw  [fill={rgb, 255:red, 0; green, 0; blue, 0 }  ,fill opacity=1 ] (466.38,328.8) -- (466.34,328.77) .. controls (466.28,329.03) and (466.17,329.27) .. (466.02,329.48) .. controls (465.21,330.58) and (463.4,330.38) .. (461.98,329.03) .. controls (460.55,327.67) and (460.06,325.68) .. (460.87,324.58) .. controls (461.05,324.32) and (461.29,324.14) .. (461.56,324.02) -- (488.73,292.61) -- (466.38,328.8) -- cycle (462.38,325.87) .. controls (461.99,326.4) and (462.24,327.35) .. (462.93,328.01) .. controls (463.62,328.66) and (464.49,328.77) .. (464.87,328.25) .. controls (465.26,327.72) and (465.01,326.77) .. (464.32,326.11) .. controls (463.63,325.46) and (462.76,325.35) .. (462.38,325.87) -- cycle ;

\draw    (413.75,518.33) -- (439.58,518.63) ;
\draw [shift={(442.58,518.67)}, rotate = 180.66] [fill={rgb, 255:red, 0; green, 0; blue, 0 }  ][line width=0.08]  [draw opacity=0] (8.93,-4.29) -- (0,0) -- (8.93,4.29) -- cycle    ;
\draw    (413.75,512) -- (439.58,512.3) ;
\draw [shift={(442.58,512.33)}, rotate = 180.66] [fill={rgb, 255:red, 0; green, 0; blue, 0 }  ][line width=0.08]  [draw opacity=0] (8.93,-4.29) -- (0,0) -- (8.93,4.29) -- cycle    ;
\draw    (413.75,504) -- (439.58,504.3) ;
\draw [shift={(442.58,504.33)}, rotate = 180.66] [fill={rgb, 255:red, 0; green, 0; blue, 0 }  ][line width=0.08]  [draw opacity=0] (8.93,-4.29) -- (0,0) -- (8.93,4.29) -- cycle    ;
\draw    (413.75,547.33) -- (439.58,547.63) ;
\draw [shift={(442.58,547.67)}, rotate = 180.66] [fill={rgb, 255:red, 0; green, 0; blue, 0 }  ][line width=0.08]  [draw opacity=0] (8.93,-4.29) -- (0,0) -- (8.93,4.29) -- cycle    ;
\draw    (413.75,541) -- (439.58,541.3) ;
\draw [shift={(442.58,541.33)}, rotate = 180.66] [fill={rgb, 255:red, 0; green, 0; blue, 0 }  ][line width=0.08]  [draw opacity=0] (8.93,-4.29) -- (0,0) -- (8.93,4.29) -- cycle    ;
\draw    (413.75,533) -- (439.58,533.3) ;
\draw [shift={(442.58,533.33)}, rotate = 180.66] [fill={rgb, 255:red, 0; green, 0; blue, 0 }  ][line width=0.08]  [draw opacity=0] (8.93,-4.29) -- (0,0) -- (8.93,4.29) -- cycle    ;
\draw    (365.11,587.88) -- (365.43,718.55) ;
\draw [shift={(365.44,721.55)}, rotate = 269.86] [fill={rgb, 255:red, 0; green, 0; blue, 0 }  ][line width=0.08]  [draw opacity=0] (8.93,-4.29) -- (0,0) -- (8.93,4.29) -- cycle    ;
\draw    (349.08,588.33) -- (349.41,719) ;
\draw [shift={(349.42,722)}, rotate = 269.86] [fill={rgb, 255:red, 0; green, 0; blue, 0 }  ][line width=0.08]  [draw opacity=0] (8.93,-4.29) -- (0,0) -- (8.93,4.29) -- cycle    ;
\draw  [color={rgb, 255:red, 74; green, 74; blue, 74 }  ,draw opacity=1 ][fill={rgb, 255:red, 255; green, 255; blue, 255 }  ,fill opacity=1 ][line width=1.5]  (194.08,446.04) .. controls (194.08,439.58) and (199.33,434.33) .. (205.79,434.33) -- (379.71,434.33) .. controls (386.18,434.33) and (391.42,439.58) .. (391.42,446.04) -- (391.42,600) .. controls (391.42,606.46) and (386.18,611.71) .. (379.71,611.71) -- (205.79,611.71) .. controls (199.33,611.71) and (194.08,606.46) .. (194.08,600) -- cycle ;
\draw  [color={rgb, 255:red, 74; green, 74; blue, 74 }  ,draw opacity=1 ][fill={rgb, 255:red, 255; green, 255; blue, 255 }  ,fill opacity=1 ][line width=1.5]  (205.79,457.75) .. controls (205.79,451.28) and (211.03,446.04) .. (217.5,446.04) -- (391.42,446.04) .. controls (397.88,446.04) and (403.13,451.28) .. (403.13,457.75) -- (403.13,611.71) .. controls (403.13,618.17) and (397.88,623.41) .. (391.42,623.41) -- (217.5,623.41) .. controls (211.03,623.41) and (205.79,618.17) .. (205.79,611.71) -- cycle ;
\draw  [color={rgb, 255:red, 74; green, 74; blue, 74 }  ,draw opacity=1 ][fill={rgb, 255:red, 255; green, 255; blue, 255 }  ,fill opacity=1 ][line width=1.5]  (217.5,469.46) .. controls (217.5,462.99) and (222.74,457.75) .. (229.21,457.75) -- (403.13,457.75) .. controls (409.59,457.75) and (414.83,462.99) .. (414.83,469.46) -- (414.83,623.41) .. controls (414.83,629.88) and (409.59,635.12) .. (403.13,635.12) -- (229.21,635.12) .. controls (222.74,635.12) and (217.5,629.88) .. (217.5,623.41) -- cycle ;
\draw   (78.33,442.33) -- (137.75,442.33) -- (137.75,673.67) -- (78.33,673.67) -- cycle ;
\draw   (111,722.33) -- (531.42,722.33) -- (531.42,784) -- (111,784) -- cycle ;
\draw    (50.08,753.67) -- (108,754.62) ;
\draw [shift={(111,754.67)}, rotate = 180.94] [fill={rgb, 255:red, 0; green, 0; blue, 0 }  ][line width=0.08]  [draw opacity=0] (8.93,-4.29) -- (0,0) -- (8.93,4.29) -- cycle    ;
\draw    (387.54,635.33) -- (387.74,719.67) ;
\draw [shift={(387.75,722.67)}, rotate = 269.86] [fill={rgb, 255:red, 0; green, 0; blue, 0 }  ][line width=0.08]  [draw opacity=0] (8.93,-4.29) -- (0,0) -- (8.93,4.29) -- cycle    ;
\draw    (47.58,572.17) -- (75.33,572.32) ;
\draw [shift={(78.33,572.33)}, rotate = 180.31] [fill={rgb, 255:red, 0; green, 0; blue, 0 }  ][line width=0.08]  [draw opacity=0] (8.93,-4.29) -- (0,0) -- (8.93,4.29) -- cycle    ;
\draw  [fill={rgb, 255:red, 80; green, 227; blue, 194 }  ,fill opacity=1 ] (51.51,519.91) -- (51.51,616.04) .. controls (51.51,623.55) and (42.56,629.64) .. (31.51,629.64) .. controls (20.47,629.64) and (11.51,623.55) .. (11.51,616.04) -- (11.51,519.91) .. controls (11.51,512.4) and (20.47,506.31) .. (31.51,506.31) .. controls (42.56,506.31) and (51.51,512.4) .. (51.51,519.91) .. controls (51.51,527.42) and (42.56,533.51) .. (31.51,533.51) .. controls (20.47,533.51) and (11.51,527.42) .. (11.51,519.91) ;
\draw  [fill={rgb, 255:red, 80; green, 227; blue, 194 }  ,fill opacity=1 ] (262.26,569.88) -- (262.26,594.88) .. controls (262.26,601.78) and (253.31,607.38) .. (242.26,607.38) .. controls (231.22,607.38) and (222.26,601.78) .. (222.26,594.88) -- (222.26,569.88) .. controls (222.26,562.98) and (231.22,557.38) .. (242.26,557.38) .. controls (253.31,557.38) and (262.26,562.98) .. (262.26,569.88) .. controls (262.26,576.78) and (253.31,582.38) .. (242.26,582.38) .. controls (231.22,582.38) and (222.26,576.78) .. (222.26,569.88) ;
\draw  [fill={rgb, 255:red, 80; green, 227; blue, 194 }  ,fill opacity=1 ] (267.93,475.83) -- (267.93,500.83) .. controls (267.93,507.74) and (258.98,513.33) .. (247.93,513.33) .. controls (236.88,513.33) and (227.93,507.74) .. (227.93,500.83) -- (227.93,475.83) .. controls (227.93,468.93) and (236.88,463.33) .. (247.93,463.33) .. controls (258.98,463.33) and (267.93,468.93) .. (267.93,475.83) .. controls (267.93,482.74) and (258.98,488.33) .. (247.93,488.33) .. controls (236.88,488.33) and (227.93,482.74) .. (227.93,475.83) ;
\draw    (220.08,519) -- (411.08,518.34) ;
\draw [shift={(414.08,518.33)}, rotate = 179.8] [fill={rgb, 255:red, 0; green, 0; blue, 0 }  ][line width=0.08]  [draw opacity=0] (8.93,-4.29) -- (0,0) -- (8.93,4.29) -- cycle    ;
\draw    (335,548) -- (410.42,548) ;
\draw [shift={(413.42,548)}, rotate = 180] [fill={rgb, 255:red, 0; green, 0; blue, 0 }  ][line width=0.08]  [draw opacity=0] (8.93,-4.29) -- (0,0) -- (8.93,4.29) -- cycle    ;
\draw   (443.7,493.62) -- (563.99,493.62) -- (563.99,573.98) -- (443.7,573.98) -- cycle ;
\draw  [color={rgb, 255:red, 74; green, 74; blue, 74 }  ,draw opacity=1 ][line width=1.5]  (66.74,405.16) .. controls (66.74,400.36) and (70.63,396.46) .. (75.43,396.46) -- (576.31,396.46) .. controls (581.11,396.46) and (585,400.36) .. (585,405.16) -- (585,796.31) .. controls (585,801.11) and (581.11,805) .. (576.31,805) -- (75.43,805) .. controls (70.63,805) and (66.74,801.11) .. (66.74,796.31) -- cycle ;
\draw    (503.51,575) -- (503.42,720) ;
\draw [shift={(503.42,723)}, rotate = 270.04] [fill={rgb, 255:red, 0; green, 0; blue, 0 }  ][line width=0.08]  [draw opacity=0] (8.93,-4.29) -- (0,0) -- (8.93,4.29) -- cycle    ;
\draw    (503.88,465) -- (503.82,489.8) ;
\draw [shift={(503.81,492.8)}, rotate = 270.14] [fill={rgb, 255:red, 0; green, 0; blue, 0 }  ][line width=0.08]  [draw opacity=0] (8.93,-4.29) -- (0,0) -- (8.93,4.29) -- cycle    ;
\draw    (531.42,751.33) -- (675.08,751.33) ;
\draw [shift={(678.08,751.33)}, rotate = 180] [fill={rgb, 255:red, 0; green, 0; blue, 0 }  ][line width=0.08]  [draw opacity=0] (8.93,-4.29) -- (0,0) -- (8.93,4.29) -- cycle    ;
\draw  [fill={rgb, 255:red, 80; green, 227; blue, 194 }  ,fill opacity=1 ] (182.18,454.14) -- (182.18,479.14) .. controls (182.18,486.05) and (173.23,491.64) .. (162.18,491.64) .. controls (151.13,491.64) and (142.18,486.05) .. (142.18,479.14) -- (142.18,454.14) .. controls (142.18,447.24) and (151.13,441.64) .. (162.18,441.64) .. controls (173.23,441.64) and (182.18,447.24) .. (182.18,454.14) .. controls (182.18,461.05) and (173.23,466.64) .. (162.18,466.64) .. controls (151.13,466.64) and (142.18,461.05) .. (142.18,454.14) ;
\draw  [fill={rgb, 255:red, 80; green, 227; blue, 194 }  ,fill opacity=1 ] (182.18,521.48) -- (182.18,546.48) .. controls (182.18,553.38) and (173.23,558.98) .. (162.18,558.98) .. controls (151.13,558.98) and (142.18,553.38) .. (142.18,546.48) -- (142.18,521.48) .. controls (142.18,514.57) and (151.13,508.98) .. (162.18,508.98) .. controls (173.23,508.98) and (182.18,514.57) .. (182.18,521.48) .. controls (182.18,528.38) and (173.23,533.98) .. (162.18,533.98) .. controls (151.13,533.98) and (142.18,528.38) .. (142.18,521.48) ;
\draw  [fill={rgb, 255:red, 80; green, 227; blue, 194 }  ,fill opacity=1 ] (182.18,614.81) -- (182.18,639.81) .. controls (182.18,646.71) and (173.23,652.31) .. (162.18,652.31) .. controls (151.13,652.31) and (142.18,646.71) .. (142.18,639.81) -- (142.18,614.81) .. controls (142.18,607.91) and (151.13,602.31) .. (162.18,602.31) .. controls (173.23,602.31) and (182.18,607.91) .. (182.18,614.81) .. controls (182.18,621.71) and (173.23,627.31) .. (162.18,627.31) .. controls (151.13,627.31) and (142.18,621.71) .. (142.18,614.81) ;
\draw  [fill={rgb, 255:red, 80; green, 227; blue, 194 }  ,fill opacity=1 ] (272.26,580.88) -- (272.26,605.88) .. controls (272.26,612.78) and (263.31,618.38) .. (252.26,618.38) .. controls (241.22,618.38) and (232.26,612.78) .. (232.26,605.88) -- (232.26,580.88) .. controls (232.26,573.98) and (241.22,568.38) .. (252.26,568.38) .. controls (263.31,568.38) and (272.26,573.98) .. (272.26,580.88) .. controls (272.26,587.78) and (263.31,593.38) .. (252.26,593.38) .. controls (241.22,593.38) and (232.26,587.78) .. (232.26,580.88) ;
\draw  [fill={rgb, 255:red, 80; green, 227; blue, 194 }  ,fill opacity=1 ] (281.26,590.88) -- (281.26,615.88) .. controls (281.26,622.78) and (272.31,628.38) .. (261.26,628.38) .. controls (250.22,628.38) and (241.26,622.78) .. (241.26,615.88) -- (241.26,590.88) .. controls (241.26,583.98) and (250.22,578.38) .. (261.26,578.38) .. controls (272.31,578.38) and (281.26,583.98) .. (281.26,590.88) .. controls (281.26,597.78) and (272.31,603.38) .. (261.26,603.38) .. controls (250.22,603.38) and (241.26,597.78) .. (241.26,590.88) ;
\draw    (219.86,547.56) -- (259.33,547.35) ;
\draw [shift={(262.33,547.33)}, rotate = 179.69] [fill={rgb, 255:red, 0; green, 0; blue, 0 }  ][line width=0.08]  [draw opacity=0] (8.93,-4.29) -- (0,0) -- (8.93,4.29) -- cycle    ;
\draw    (137.25,566) -- (163.08,566.3) ;
\draw [shift={(166.08,566.33)}, rotate = 180.66] [fill={rgb, 255:red, 0; green, 0; blue, 0 }  ][line width=0.08]  [draw opacity=0] (8.93,-4.29) -- (0,0) -- (8.93,4.29) -- cycle    ;
\draw    (137.92,496) -- (163.75,496.3) ;
\draw [shift={(166.75,496.33)}, rotate = 180.66] [fill={rgb, 255:red, 0; green, 0; blue, 0 }  ][line width=0.08]  [draw opacity=0] (8.93,-4.29) -- (0,0) -- (8.93,4.29) -- cycle    ;
\draw    (139.25,658.67) -- (165.08,658.97) ;
\draw [shift={(168.08,659)}, rotate = 180.66] [fill={rgb, 255:red, 0; green, 0; blue, 0 }  ][line width=0.08]  [draw opacity=0] (8.93,-4.29) -- (0,0) -- (8.93,4.29) -- cycle    ;
\draw  [color={rgb, 255:red, 255; green, 255; blue, 255 }  ,draw opacity=1 ][fill={rgb, 255:red, 126; green, 211; blue, 33 }  ,fill opacity=1 ] (444.37,749.88) .. controls (440.61,752.02) and (437.22,754.72) .. (434.33,757.87) -- (429.47,754.75) .. controls (432.28,750.56) and (435.76,746.83) .. (439.74,743.71) -- (444.37,749.88) -- cycle ;
\draw  [color={rgb, 255:red, 255; green, 255; blue, 255 }  ,draw opacity=1 ][fill={rgb, 255:red, 184; green, 233; blue, 134 }  ,fill opacity=1 ] (453.93,746.32) .. controls (450.34,747.09) and (446.94,748.38) .. (443.81,750.1) -- (439.43,744.15) .. controls (443,741.32) and (447.04,739.04) .. (451.4,737.43) -- (453.93,746.32) -- cycle ;
\draw  [color={rgb, 255:red, 255; green, 255; blue, 255 }  ,draw opacity=1 ][fill={rgb, 255:red, 248; green, 217; blue, 166 }  ,fill opacity=1 ] (465.18,745.47) .. controls (464.37,745.42) and (463.54,745.39) .. (462.71,745.39) .. controls (459.69,745.39) and (456.75,745.71) .. (453.93,746.32) -- (451.44,737.7) .. controls (456.05,736.09) and (461.01,735.16) .. (466.18,735.03) -- (465.18,745.47) -- cycle ;
\draw  [color={rgb, 255:red, 255; green, 255; blue, 255 }  ,draw opacity=1 ][fill={rgb, 255:red, 248; green, 231; blue, 28 }  ,fill opacity=1 ] (467.12,734.75) .. controls (472.31,734.75) and (477.3,735.6) .. (481.96,737.18) -- (476.81,747.78) .. controls (473.21,746.41) and (469.35,745.55) .. (465.34,745.29) -- (466.35,734.75) .. controls (466.61,734.75) and (466.87,734.75) .. (467.12,734.75) -- cycle ;
\draw  [color={rgb, 255:red, 255; green, 255; blue, 255 }  ,draw opacity=1 ][fill={rgb, 255:red, 245; green, 166; blue, 35 }  ,fill opacity=1 ] (486.47,752.81) .. controls (483.48,750.66) and (480.13,748.89) .. (476.53,747.6) -- (481.96,737.18) .. controls (487.12,738.81) and (491.87,741.25) .. (496.02,744.34) -- (486.47,752.81) -- cycle ;
\draw  [color={rgb, 255:red, 255; green, 255; blue, 255 }  ,draw opacity=1 ][fill={rgb, 255:red, 208; green, 2; blue, 27 }  ,fill opacity=1 ] (496.11,763.43) .. controls (493.49,759.42) and (490.15,755.83) .. (486.26,752.81) -- (495.88,744.54) .. controls (501.12,748.4) and (505.56,753.12) .. (508.93,758.47) -- (496.11,763.43) -- cycle ;
\draw  [color={rgb, 255:red, 255; green, 255; blue, 255 }  ,draw opacity=1 ][fill={rgb, 255:red, 65; green, 117; blue, 5 }  ,fill opacity=1 ][line width=0.75]  (434.43,758) .. controls (431.54,761.17) and (429.15,764.8) .. (427.41,768.79) -- (423.48,767.89) .. controls (424.78,763.16) and (426.83,758.74) .. (429.47,754.75) -- (434.43,758) -- cycle ;
\draw  [draw opacity=0][line width=2.25]  (420.11,767.08) .. controls (424.92,747.84) and (443.52,733.21) .. (466.11,732.63) .. controls (485.81,732.12) and (503.03,742.43) .. (510.96,757.66) -- (467.25,776.74) -- cycle ; \draw  [color={rgb, 255:red, 155; green, 155; blue, 155 }  ,draw opacity=1 ][line width=2.25]  (420.11,767.08) .. controls (424.92,747.84) and (443.52,733.21) .. (466.11,732.63) .. controls (485.81,732.12) and (503.03,742.43) .. (510.96,757.66) ;  
\draw  [fill={rgb, 255:red, 0; green, 0; blue, 0 }  ,fill opacity=1 ] (466.71,780.47) -- (466.67,780.44) .. controls (466.61,780.7) and (466.51,780.93) .. (466.35,781.14) .. controls (465.54,782.25) and (463.73,782.05) .. (462.31,780.69) .. controls (460.89,779.34) and (460.39,777.35) .. (461.2,776.24) .. controls (461.39,775.99) and (461.62,775.81) .. (461.89,775.69) -- (489.07,744.27) -- (466.71,780.47) -- cycle (462.71,777.54) .. controls (462.33,778.06) and (462.57,779.02) .. (463.26,779.67) .. controls (463.95,780.33) and (464.82,780.44) .. (465.21,779.91) .. controls (465.59,779.39) and (465.34,778.43) .. (464.65,777.78) .. controls (463.96,777.12) and (463.09,777.02) .. (462.71,777.54) -- cycle ;

\draw   (264.33,530.57) -- (335,530.57) -- (335,565.21) -- (264.33,565.21) -- cycle ;

\draw   (443.37,123) -- (563.66,123) -- (563.66,179) -- (443.37,179) -- cycle ;
\draw  [fill={rgb, 255:red, 245; green, 166; blue, 35 }  ,fill opacity=1 ] (302.29,587.44) .. controls (302.29,581.75) and (306.91,577.13) .. (312.6,577.13) -- (373.36,577.13) .. controls (379.05,577.13) and (383.67,581.75) .. (383.67,587.44) -- (383.67,618.36) .. controls (383.67,624.05) and (379.05,628.67) .. (373.36,628.67) -- (312.6,628.67) .. controls (306.91,628.67) and (302.29,624.05) .. (302.29,618.36) -- cycle ;
\draw [fill={rgb, 255:red, 255; green, 255; blue, 255 }  ,fill opacity=1 ]   (309.52,594.41) -- (319.63,594.41) ;
\draw [fill={rgb, 255:red, 255; green, 255; blue, 255 }  ,fill opacity=1 ]   (309.66,611.5) -- (319.77,611.5) ;
\draw [fill={rgb, 255:red, 255; green, 255; blue, 255 }  ,fill opacity=1 ]   (338.48,585.32) -- (357.84,594.39) ;
\draw [fill={rgb, 255:red, 255; green, 255; blue, 255 }  ,fill opacity=1 ]   (338.48,585.32) -- (357.84,611.63) ;
\draw [fill={rgb, 255:red, 255; green, 255; blue, 255 }  ,fill opacity=1 ]   (338.48,602.44) -- (357.84,611.63) ;
\draw [fill={rgb, 255:red, 255; green, 255; blue, 255 }  ,fill opacity=1 ]   (338.48,619.81) -- (357.84,594.39) ;
\draw [fill={rgb, 255:red, 255; green, 255; blue, 255 }  ,fill opacity=1 ]   (338.48,602.57) -- (357.84,594.39) ;
\draw [fill={rgb, 255:red, 255; green, 255; blue, 255 }  ,fill opacity=1 ]   (338.48,619.81) -- (357.84,611.63) ;
\draw  [fill={rgb, 255:red, 255; green, 255; blue, 255 }  ,fill opacity=1 ] (353.54,594.39) .. controls (353.54,592.09) and (355.46,590.22) .. (357.84,590.22) .. controls (360.22,590.22) and (362.15,592.09) .. (362.15,594.39) .. controls (362.15,596.69) and (360.22,598.55) .. (357.84,598.55) .. controls (355.46,598.55) and (353.54,596.69) .. (353.54,594.39) -- cycle ;
\draw  [fill={rgb, 255:red, 255; green, 255; blue, 255 }  ,fill opacity=1 ] (353.54,611.63) .. controls (353.54,609.33) and (355.46,607.47) .. (357.84,607.47) .. controls (360.22,607.47) and (362.15,609.33) .. (362.15,611.63) .. controls (362.15,613.93) and (360.22,615.79) .. (357.84,615.79) .. controls (355.46,615.79) and (353.54,613.93) .. (353.54,611.63) -- cycle ;
\draw [fill={rgb, 255:red, 255; green, 255; blue, 255 }  ,fill opacity=1 ]   (362.15,594.39) -- (378.34,594.39) ;
\draw [fill={rgb, 255:red, 255; green, 255; blue, 255 }  ,fill opacity=1 ]   (362.15,611.63) -- (378.34,611.63) ;
\draw  [fill={rgb, 255:red, 255; green, 255; blue, 255 }  ,fill opacity=1 ] (366.09,585.07) -- (373.23,585.07) -- (373.23,618.57) -- (366.09,618.57) -- cycle ;
\draw [fill={rgb, 255:red, 255; green, 255; blue, 255 }  ,fill opacity=1 ]   (338.5,585.48) -- (319.14,594.54) ;
\draw [fill={rgb, 255:red, 255; green, 255; blue, 255 }  ,fill opacity=1 ]   (338.5,585.48) -- (319.14,611.79) ;
\draw [fill={rgb, 255:red, 255; green, 255; blue, 255 }  ,fill opacity=1 ]   (338.5,602.6) -- (319.14,611.79) ;
\draw [fill={rgb, 255:red, 255; green, 255; blue, 255 }  ,fill opacity=1 ]   (338.5,619.97) -- (319.14,594.54) ;
\draw [fill={rgb, 255:red, 255; green, 255; blue, 255 }  ,fill opacity=1 ]   (338.5,602.72) -- (319.14,594.54) ;
\draw [fill={rgb, 255:red, 255; green, 255; blue, 255 }  ,fill opacity=1 ]   (338.5,619.97) -- (319.14,611.79) ;
\draw  [fill={rgb, 255:red, 255; green, 255; blue, 255 }  ,fill opacity=1 ] (315.56,594.39) .. controls (315.56,592.09) and (317.49,590.22) .. (319.86,590.22) .. controls (322.24,590.22) and (324.17,592.09) .. (324.17,594.39) .. controls (324.17,596.69) and (322.24,598.55) .. (319.86,598.55) .. controls (317.49,598.55) and (315.56,596.69) .. (315.56,594.39) -- cycle ;
\draw  [fill={rgb, 255:red, 255; green, 255; blue, 255 }  ,fill opacity=1 ] (315.56,611.63) .. controls (315.56,609.33) and (317.49,607.47) .. (319.86,607.47) .. controls (322.24,607.47) and (324.17,609.33) .. (324.17,611.63) .. controls (324.17,613.93) and (322.24,615.79) .. (319.86,615.79) .. controls (317.49,615.79) and (315.56,613.93) .. (315.56,611.63) -- cycle ;
\draw  [fill={rgb, 255:red, 255; green, 255; blue, 255 }  ,fill opacity=1 ] (334.17,585.32) .. controls (334.17,583.03) and (336.1,581.16) .. (338.48,581.16) .. controls (340.85,581.16) and (342.78,583.03) .. (342.78,585.32) .. controls (342.78,587.62) and (340.85,589.49) .. (338.48,589.49) .. controls (336.1,589.49) and (334.17,587.62) .. (334.17,585.32) -- cycle ;
\draw  [fill={rgb, 255:red, 255; green, 255; blue, 255 }  ,fill opacity=1 ] (334.17,602.57) .. controls (334.17,600.27) and (336.1,598.4) .. (338.48,598.4) .. controls (340.85,598.4) and (342.78,600.27) .. (342.78,602.57) .. controls (342.78,604.87) and (340.85,606.73) .. (338.48,606.73) .. controls (336.1,606.73) and (334.17,604.87) .. (334.17,602.57) -- cycle ;
\draw  [fill={rgb, 255:red, 255; green, 255; blue, 255 }  ,fill opacity=1 ] (334.17,619.81) .. controls (334.17,617.51) and (336.1,615.65) .. (338.48,615.65) .. controls (340.85,615.65) and (342.78,617.51) .. (342.78,619.81) .. controls (342.78,622.11) and (340.85,623.97) .. (338.48,623.97) .. controls (336.1,623.97) and (334.17,622.11) .. (334.17,619.81) -- cycle ;

\draw    (387.33,549) -- (387.53,632.33) ;
\draw [shift={(387.54,635.33)}, rotate = 269.86] [fill={rgb, 255:red, 0; green, 0; blue, 0 }  ][line width=0.08]  [draw opacity=0] (8.93,-4.29) -- (0,0) -- (8.93,4.29) -- cycle    ;

\draw (177.11,220.04) node  [font=\Large]  {$\mathcal{D}^{\text{tr}}$};
\draw (179.02,106.37) node  [font=\LARGE]  {$\mathcal{D}^{\text{val}}$};
\draw (31.91,150.45) node  [font=\LARGE]  {$\mathcal{D}$};
\draw (276.92,304.12) node  [font=\LARGE]  {$\text{predictive set} \ \ \ \Gamma _{\lambda }\left( x|\mathcal{D}^{\text{tr}}\right) \ $};
\draw (590.78,276.5) node [anchor=west] [inner sep=0.75pt]  [font=\LARGE]  {$\Gamma ^{\text{VB}}( x|\mathcal{D}) \ $};
\draw (499.73,76.58) node  [font=\LARGE]  {$\text{target risk } \alpha $};
\draw (59.83,286.83) node [anchor=east] [inner sep=0.75pt]  [font=\LARGE] [align=left] {\begin{minipage}[lt]{45.9pt}\setlength\topsep{0pt}
\begin{flushright}
test\\input \\$\displaystyle x$
\end{flushright}

\end{minipage}};
\draw (505.8,149.14) node  [font=\normalsize] [align=left] {\begin{minipage}[lt]{88.04pt}\setlength\topsep{0pt}
\begin{center}
{\LARGE VB}\\{\LARGE risk control}
\end{center}

\end{minipage}};
\draw (106.4,147.99) node  [font=\LARGE] [align=left] {split};
\draw (457.68,227.06) node  [font=\Large]  {$ \begin{array}{l}
\text{threshold}\\
\lambda \left(\mathcal{D}^{\text{val}} |\mathcal{D}^{\text{tr}}\right)
\end{array}$};
\draw (260.2,611.95) node  [font=\LARGE]  {$\mathcal{D}_{-k}$};
\draw (249.8,498.59) node  [font=\LARGE]  {$\mathcal{D}_{k}$};
\draw (259.19,751.3) node  [font=\LARGE]  {$\text{predictive set} \ \bigcup _{k=1}^{K} \Gamma _{\lambda }( x|\mathcal{D}_{-k}) \ $};
\draw (495.4,665.34) node  [font=\Large]  {$\text{threshold} \ \lambda ^{\text{CV}}(\mathcal{D})$};
\draw (588.71,726.76) node [anchor=west] [inner sep=0.75pt]  [font=\LARGE]  {$\Gamma ^{\text{CV}}( x|\mathcal{D}) \ $};
\draw (505.53,449.19) node  [font=\LARGE]  {$\text{target risk } \alpha $};
\draw (504.32,533.33) node  [font=\LARGE] [align=left] {\begin{minipage}[lt]{88.04pt}\setlength\topsep{0pt}
\begin{center}
CV \\risk control
\end{center}

\end{minipage}};
\draw (110.15,559.99) node  [font=\LARGE] [align=left] {$\displaystyle K$-\\fold\\split};
\draw (164.37,476.9) node  [font=\LARGE]  {$\mathcal{D}_{1}$};
\draw (164.37,544.24) node  [font=\LARGE]  {$\mathcal{D}_{2}$};
\draw (166.09,637.57) node  [font=\LARGE]  {$\mathcal{D}_{K}$};
\draw (148.85,578.67) node [anchor=north west][inner sep=0.75pt]   [align=left] {...};
\draw (357.3,472.35) node  [font=\LARGE]  {$k=1,...,K$};
\draw (374.85,685.44) node  [font=\Large] [align=left] {...};
\draw (257.5,117.23) node  [font=\LARGE] [align=left] {validation};
\draw (255,163.56) node  [font=\LARGE] [align=left] {train};
\draw (31.91,570.33) node  [font=\LARGE]  {$\mathcal{D}$};
\draw (61.17,736.04) node [anchor=east] [inner sep=0.75pt]  [font=\LARGE] [align=left] {\begin{minipage}[lt]{45.9pt}\setlength\topsep{0pt}
\begin{flushright}
test\\input \\$\displaystyle x$
\end{flushright}

\end{minipage}};
\draw (411.9,303.43) node  [font=\Large]  {$\emptyset $};
\draw (515.98,290.65) node  [font=\Large]  {$\mathcal{Y}$};
\draw (457.39,308.43) node  [font=\Large]  {$\lambda $};
\draw (457.72,760.1) node  [font=\Large]  {$\lambda $};
\draw (516.32,742.32) node  [font=\Large]  {$\mathcal{Y}$};
\draw (412.23,755.1) node  [font=\Large]  {$\emptyset $};
\draw (299.67,547.89) node  [font=\LARGE] [align=left] {train};
\draw (338.83,507.56) node  [font=\LARGE] [align=left] {\begin{minipage}[lt]{59.62pt}\setlength\topsep{0pt}
\begin{flushright}
fold set
\end{flushright}

\end{minipage}};

\end{tikzpicture}



%% file: Figs/fig_tikz_point_process.tex
\tikzset{every picture/.style={line width=0.75pt}} 

\begin{tikzpicture}[x=0.75pt,y=0.75pt,yscale=-1,xscale=1]

\draw  [draw opacity=0][fill={rgb, 255:red, 255; green, 214; blue, 214 }  ,fill opacity=1 ] (326,59.43) -- (495.67,59.43) -- (495.67,220.67) -- (326,220.67) -- cycle ;
\draw  [fill={rgb, 255:red, 190; green, 255; blue, 240 }  ,fill opacity=1 ] (75.8,125.32) -- (75.8,184.01) .. controls (75.8,187.13) and (60.25,189.67) .. (41.07,189.67) .. controls (21.88,189.67) and (6.33,187.13) .. (6.33,184.01) -- (6.33,125.32) .. controls (6.33,122.2) and (21.88,119.67) .. (41.07,119.67) .. controls (60.25,119.67) and (75.8,122.2) .. (75.8,125.32) .. controls (75.8,128.45) and (60.25,130.98) .. (41.07,130.98) .. controls (21.88,130.98) and (6.33,128.45) .. (6.33,125.32) ;
\draw [color={rgb, 255:red, 74; green, 144; blue, 226 }  ,draw opacity=1 ]   (5.8,44.73) -- (5.8,74.13) ;
\draw [color={rgb, 255:red, 74; green, 144; blue, 226 }  ,draw opacity=1 ]   (53,44.73) -- (53,74.13) ;
\draw [color={rgb, 255:red, 74; green, 144; blue, 226 }  ,draw opacity=1 ]   (94.77,44.73) -- (94.77,74.13) ;
\draw [color={rgb, 255:red, 74; green, 144; blue, 226 }  ,draw opacity=1 ]   (122.66,44.73) -- (122.66,74.13) ;
\draw [color={rgb, 255:red, 74; green, 144; blue, 226 }  ,draw opacity=1 ]   (132.86,44.73) -- (132.86,74.13) ;
\draw [color={rgb, 255:red, 74; green, 144; blue, 226 }  ,draw opacity=1 ]   (141.66,44.73) -- (141.66,74.13) ;
\draw [color={rgb, 255:red, 74; green, 144; blue, 226 }  ,draw opacity=1 ]   (149.67,44.73) -- (149.67,74.13) ;
\draw [color={rgb, 255:red, 74; green, 144; blue, 226 }  ,draw opacity=1 ]   (175.67,44.73) -- (175.67,74.13) ;
\draw [color={rgb, 255:red, 74; green, 144; blue, 226 }  ,draw opacity=1 ]   (217.67,44.73) -- (217.67,74.13) ;
\draw [color={rgb, 255:red, 74; green, 144; blue, 226 }  ,draw opacity=1 ]   (270,44.73) -- (270,74.13) ;
\draw [color={rgb, 255:red, 74; green, 144; blue, 226 }  ,draw opacity=1 ][line width=1.5]    (345.44,44.73) -- (345.44,74.13) ;
\draw [color={rgb, 255:red, 74; green, 144; blue, 226 }  ,draw opacity=1 ][line width=1.5]    (399.32,44.73) -- (399.32,74.13) ;
\draw [color={rgb, 255:red, 74; green, 144; blue, 226 }  ,draw opacity=1 ][line width=1.5]    (422.52,44.73) -- (422.52,74.13) ;
\draw [color={rgb, 255:red, 74; green, 144; blue, 226 }  ,draw opacity=1 ][line width=1.5]    (435.32,44.73) -- (435.32,74.13) ;
\draw [color={rgb, 255:red, 74; green, 144; blue, 226 }  ,draw opacity=1 ][line width=1.5]    (443.33,44.73) -- (443.33,74.13) ;
\draw [color={rgb, 255:red, 74; green, 144; blue, 226 }  ,draw opacity=1 ][line width=1.5]    (458.33,44.73) -- (458.33,74.13) ;
\draw  [line width=0.75] [line join = round][line cap = round] (402.57,25.66) .. controls (392.9,49.69) and (354.18,9.36) .. (344.03,39.77) ;
\draw  [line width=0.75] [line join = round][line cap = round] (403.43,25.66) .. controls (413.1,49.69) and (451.82,9.36) .. (461.97,39.77) ;
\draw  [line width=0.75] [line join = round][line cap = round] (165.7,114.11) .. controls (139.44,90.07) and (34.35,130.4) .. (6.79,99.99) ;
\draw  [line width=0.75] [line join = round][line cap = round] (165.64,114.77) .. controls (191.9,90.74) and (296.98,131.07) .. (324.55,100.66) ;
\draw  [draw opacity=0][fill={rgb, 255:red, 126; green, 211; blue, 33 }  ,fill opacity=1 ] (338.2,125) -- (352.4,125) -- (352.4,141) -- (338.2,141) -- cycle ;
\draw  [draw opacity=0][fill={rgb, 255:red, 126; green, 211; blue, 33 }  ,fill opacity=1 ] (359.2,141) -- (410.4,141) -- (410.4,157) -- (359.2,157) -- cycle ;
\draw  [draw opacity=0][fill={rgb, 255:red, 126; green, 211; blue, 33 }  ,fill opacity=1 ] (408,156.4) -- (427,156.4) -- (427,172.4) -- (408,172.4) -- cycle ;
\draw  [draw opacity=0][fill={rgb, 255:red, 126; green, 211; blue, 33 }  ,fill opacity=1 ] (411,172.4) -- (430,172.4) -- (430,188.4) -- (411,188.4) -- cycle ;
\draw  [draw opacity=0][fill={rgb, 255:red, 126; green, 211; blue, 33 }  ,fill opacity=1 ] (420.8,188.4) -- (470.8,188.4) -- (470.8,204.4) -- (420.8,204.4) -- cycle ;
\draw  [draw opacity=0][fill={rgb, 255:red, 126; green, 211; blue, 33 }  ,fill opacity=1 ] (455,204.4) -- (493.6,204.4) -- (493.6,220.4) -- (455,220.4) -- cycle ;
\draw [color={rgb, 255:red, 74; green, 144; blue, 226 }  ,draw opacity=1 ][line width=1.5]    (345.44,124.73) -- (345.44,140.8) ;
\draw [color={rgb, 255:red, 74; green, 144; blue, 226 }  ,draw opacity=1 ][line width=1.5]    (399.32,140.73) -- (399.32,156.8) ;
\draw [color={rgb, 255:red, 74; green, 144; blue, 226 }  ,draw opacity=1 ][line width=1.5]    (422.52,156.73) -- (422.52,172.8) ;
\draw [color={rgb, 255:red, 208; green, 2; blue, 27 }  ,draw opacity=1 ][line width=1.5]    (435.32,172.73) -- (435.32,188.8) ;
\draw [color={rgb, 255:red, 74; green, 144; blue, 226 }  ,draw opacity=1 ][line width=1.5]    (443.33,187.73) -- (443.33,203.8) ;
\draw [color={rgb, 255:red, 74; green, 144; blue, 226 }  ,draw opacity=1 ][line width=1.5]    (458.33,204.73) -- (458.33,220.8) ;
\draw  [fill={rgb, 255:red, 255; green, 220; blue, 157 }  ,fill opacity=1 ] (107,140.81) -- (222.67,140.81) -- (222.67,209) -- (107,209) -- cycle ;
\draw [color={rgb, 255:red, 74; green, 144; blue, 226 }  ,draw opacity=1 ][line width=0.75]    (13.07,137.67) -- (13.07,148.73) ;
\draw [color={rgb, 255:red, 74; green, 144; blue, 226 }  ,draw opacity=1 ][line width=0.75]    (11.45,137.67) -- (11.45,148.73) ;
\draw [color={rgb, 255:red, 74; green, 144; blue, 226 }  ,draw opacity=1 ][line width=0.75]    (14.06,137.67) -- (14.06,148.73) ;
\draw [color={rgb, 255:red, 74; green, 144; blue, 226 }  ,draw opacity=1 ][line width=0.75]    (23.28,137.67) -- (23.28,148.73) ;
\draw [color={rgb, 255:red, 74; green, 144; blue, 226 }  ,draw opacity=1 ][line width=0.75]    (28.16,137.67) -- (28.16,148.73) ;
\draw [color={rgb, 255:red, 74; green, 144; blue, 226 }  ,draw opacity=1 ][line width=0.75]    (32.48,137.67) -- (32.48,148.73) ;
\draw [color={rgb, 255:red, 74; green, 144; blue, 226 }  ,draw opacity=1 ][line width=0.75]    (35.36,137.67) -- (35.36,148.73) ;
\draw [color={rgb, 255:red, 74; green, 144; blue, 226 }  ,draw opacity=1 ][line width=0.75]    (16.8,137.67) -- (16.8,148.73) ;
\draw [color={rgb, 255:red, 74; green, 144; blue, 226 }  ,draw opacity=1 ][line width=0.75]    (40.84,137.67) -- (40.84,148.73) ;
\draw [color={rgb, 255:red, 74; green, 144; blue, 226 }  ,draw opacity=1 ][line width=0.75]    (45.18,137.67) -- (45.18,148.73) ;
\draw [color={rgb, 255:red, 74; green, 144; blue, 226 }  ,draw opacity=1 ][line width=0.75]    (50.59,137.67) -- (50.59,148.73) ;
\draw [color={rgb, 255:red, 74; green, 144; blue, 226 }  ,draw opacity=1 ][line width=0.75]    (52.81,137.67) -- (52.81,148.73) ;
\draw [color={rgb, 255:red, 74; green, 144; blue, 226 }  ,draw opacity=1 ][line width=0.75]    (58.38,137.67) -- (58.38,148.73) ;
\draw [color={rgb, 255:red, 74; green, 144; blue, 226 }  ,draw opacity=1 ][line width=0.75]    (60.78,137.67) -- (60.78,148.73) ;
\draw [color={rgb, 255:red, 74; green, 144; blue, 226 }  ,draw opacity=1 ][line width=0.75]    (62.1,137.67) -- (62.1,148.73) ;
\draw [color={rgb, 255:red, 74; green, 144; blue, 226 }  ,draw opacity=1 ][line width=0.75]    (62.93,137.67) -- (62.93,148.73) ;
\draw [color={rgb, 255:red, 74; green, 144; blue, 226 }  ,draw opacity=1 ][line width=0.75]    (64.48,137.67) -- (64.48,148.73) ;
\draw [color={rgb, 255:red, 74; green, 144; blue, 226 }  ,draw opacity=1 ][line width=0.75]    (13.03,154.61) -- (13.03,165.67) ;
\draw [color={rgb, 255:red, 74; green, 144; blue, 226 }  ,draw opacity=1 ][line width=0.75]    (17.23,154.61) -- (17.23,165.67) ;
\draw [color={rgb, 255:red, 74; green, 144; blue, 226 }  ,draw opacity=1 ][line width=0.75]    (19.2,154.61) -- (19.2,165.67) ;
\draw [color={rgb, 255:red, 74; green, 144; blue, 226 }  ,draw opacity=1 ][line width=0.75]    (22.71,154.61) -- (22.71,165.67) ;
\draw [color={rgb, 255:red, 74; green, 144; blue, 226 }  ,draw opacity=1 ][line width=0.75]    (27.06,154.61) -- (27.06,165.67) ;
\draw [color={rgb, 255:red, 74; green, 144; blue, 226 }  ,draw opacity=1 ][line width=0.75]    (32.47,154.61) -- (32.47,165.67) ;
\draw [color={rgb, 255:red, 74; green, 144; blue, 226 }  ,draw opacity=1 ][line width=0.75]    (37.27,154.61) -- (37.27,165.67) ;
\draw [color={rgb, 255:red, 74; green, 144; blue, 226 }  ,draw opacity=1 ][line width=0.75]    (42.84,154.61) -- (42.84,165.67) ;
\draw [color={rgb, 255:red, 74; green, 144; blue, 226 }  ,draw opacity=1 ][line width=0.75]    (45.24,154.61) -- (45.24,165.67) ;
\draw [color={rgb, 255:red, 74; green, 144; blue, 226 }  ,draw opacity=1 ][line width=0.75]    (46.56,154.61) -- (46.56,165.67) ;
\draw [color={rgb, 255:red, 74; green, 144; blue, 226 }  ,draw opacity=1 ][line width=0.75]    (50.94,154.61) -- (50.94,165.67) ;
\draw [color={rgb, 255:red, 74; green, 144; blue, 226 }  ,draw opacity=1 ][line width=0.75]    (12.83,168.53) -- (12.83,179.6) ;
\draw [color={rgb, 255:red, 74; green, 144; blue, 226 }  ,draw opacity=1 ][line width=0.75]    (11.21,168.53) -- (11.21,179.6) ;
\draw [color={rgb, 255:red, 74; green, 144; blue, 226 }  ,draw opacity=1 ][line width=0.75]    (13.82,168.53) -- (13.82,179.6) ;
\draw [color={rgb, 255:red, 74; green, 144; blue, 226 }  ,draw opacity=1 ][line width=0.75]    (23.04,168.53) -- (23.04,179.6) ;
\draw [color={rgb, 255:red, 74; green, 144; blue, 226 }  ,draw opacity=1 ][line width=0.75]    (27.92,168.53) -- (27.92,179.6) ;
\draw [color={rgb, 255:red, 74; green, 144; blue, 226 }  ,draw opacity=1 ][line width=0.75]    (32.24,168.53) -- (32.24,179.6) ;
\draw [color={rgb, 255:red, 74; green, 144; blue, 226 }  ,draw opacity=1 ][line width=0.75]    (35.12,168.53) -- (35.12,179.6) ;
\draw [color={rgb, 255:red, 74; green, 144; blue, 226 }  ,draw opacity=1 ][line width=0.75]    (36.17,168.53) -- (36.17,179.6) ;
\draw [color={rgb, 255:red, 74; green, 144; blue, 226 }  ,draw opacity=1 ][line width=0.75]    (16.56,168.53) -- (16.56,179.6) ;
\draw [color={rgb, 255:red, 74; green, 144; blue, 226 }  ,draw opacity=1 ][line width=0.75]    (40.6,168.53) -- (40.6,179.6) ;
\draw [color={rgb, 255:red, 74; green, 144; blue, 226 }  ,draw opacity=1 ][line width=0.75]    (44.94,168.53) -- (44.94,179.6) ;
\draw [color={rgb, 255:red, 74; green, 144; blue, 226 }  ,draw opacity=1 ][line width=0.75]    (50.35,168.53) -- (50.35,179.6) ;
\draw [color={rgb, 255:red, 74; green, 144; blue, 226 }  ,draw opacity=1 ][line width=0.75]    (53.08,168.53) -- (53.08,179.6) ;
\draw [color={rgb, 255:red, 74; green, 144; blue, 226 }  ,draw opacity=1 ][line width=0.75]    (28.4,154.61) -- (28.4,165.67) ;
\draw [color={rgb, 255:red, 74; green, 144; blue, 226 }  ,draw opacity=1 ][line width=0.75]    (29.79,154.61) -- (29.79,165.67) ;
\draw [color={rgb, 255:red, 74; green, 144; blue, 226 }  ,draw opacity=1 ][line width=0.75]    (35.13,154.61) -- (35.13,165.67) ;
\draw [color={rgb, 255:red, 74; green, 144; blue, 226 }  ,draw opacity=1 ][line width=0.75]    (19.28,168.53) -- (19.28,179.6) ;
\draw [color={rgb, 255:red, 74; green, 144; blue, 226 }  ,draw opacity=1 ][line width=0.75]    (21.88,168.53) -- (21.88,179.6) ;
\draw [color={rgb, 255:red, 74; green, 144; blue, 226 }  ,draw opacity=1 ][line width=0.75]    (24.62,168.53) -- (24.62,179.6) ;
\draw [color={rgb, 255:red, 74; green, 144; blue, 226 }  ,draw opacity=1 ][line width=0.75]    (41.94,168.53) -- (41.94,179.6) ;
\draw [color={rgb, 255:red, 74; green, 144; blue, 226 }  ,draw opacity=1 ][line width=0.75]    (43.33,168.53) -- (43.33,179.6) ;
\draw [color={rgb, 255:red, 74; green, 144; blue, 226 }  ,draw opacity=1 ][line width=0.75]    (45.93,168.53) -- (45.93,179.6) ;
\draw [color={rgb, 255:red, 74; green, 144; blue, 226 }  ,draw opacity=1 ][line width=0.75]    (51.47,168.53) -- (51.47,179.6) ;
\draw    (9.24,148.68) -- (72.33,148.44) ;
\draw [shift={(75.33,148.43)}, rotate = 179.78] [fill={rgb, 255:red, 0; green, 0; blue, 0 }  ][line width=0.08]  [draw opacity=0] (8.93,-4.29) -- (0,0) -- (8.93,4.29) -- cycle    ;
\draw    (9.21,165.62) -- (72.3,165.38) ;
\draw [shift={(75.3,165.37)}, rotate = 179.78] [fill={rgb, 255:red, 0; green, 0; blue, 0 }  ][line width=0.08]  [draw opacity=0] (8.93,-4.29) -- (0,0) -- (8.93,4.29) -- cycle    ;
\draw    (9,179.55) -- (72.09,179.31) ;
\draw [shift={(75.09,179.3)}, rotate = 179.78] [fill={rgb, 255:red, 0; green, 0; blue, 0 }  ][line width=0.08]  [draw opacity=0] (8.93,-4.29) -- (0,0) -- (8.93,4.29) -- cycle    ;
\draw    (1,73.68) -- (513.33,75) ;
\draw [shift={(516.33,75.01)}, rotate = 180.15] [fill={rgb, 255:red, 0; green, 0; blue, 0 }  ][line width=0.08]  [draw opacity=0] (8.93,-4.29) -- (0,0) -- (8.93,4.29) -- cycle    ;
\draw    (78.6,170.07) -- (103.57,170.3) ;
\draw [shift={(105.57,170.31)}, rotate = 180.52] [color={rgb, 255:red, 0; green, 0; blue, 0 }  ][line width=0.75]    (10.93,-3.29) .. controls (6.95,-1.4) and (3.31,-0.3) .. (0,0) .. controls (3.31,0.3) and (6.95,1.4) .. (10.93,3.29)   ;
\draw    (222.33,173.68) -- (321,173.99) ;
\draw [shift={(323,174)}, rotate = 180.18] [color={rgb, 255:red, 0; green, 0; blue, 0 }  ][line width=0.75]    (10.93,-3.29) .. controls (6.95,-1.4) and (3.31,-0.3) .. (0,0) .. controls (3.31,0.3) and (6.95,1.4) .. (10.93,3.29)   ;
\draw    (78.2,200.27) -- (103.17,200.5) ;
\draw [shift={(105.17,200.51)}, rotate = 180.52] [color={rgb, 255:red, 0; green, 0; blue, 0 }  ][line width=0.75]    (10.93,-3.29) .. controls (6.95,-1.4) and (3.31,-0.3) .. (0,0) .. controls (3.31,0.3) and (6.95,1.4) .. (10.93,3.29)   ;
\draw  [line width=0.75] [line join = round][line cap = round] (345.82,148.67) .. controls (344.58,137.88) and (339.61,155.98) .. (338.3,142.33) ;
\draw  [line width=0.75] [line join = round][line cap = round] (346.15,148.67) .. controls (347.39,137.88) and (352.36,155.98) .. (353.67,142.33) ;

\draw  [line width=0.75] [line join = round][line cap = round] (473.76,229.67) .. controls (470.6,218.88) and (457.98,236.98) .. (454.67,223.33) ;
\draw  [line width=0.75] [line join = round][line cap = round] (474.58,229.67) .. controls (477.73,218.88) and (490.36,236.98) .. (493.67,223.33) ;

\draw [color={rgb, 255:red, 74; green, 144; blue, 226 }  ,draw opacity=1 ]   (326,44.73) -- (326,74.13) ;

\draw (0,78.73) node [anchor=north west][inner sep=0.75pt]    {$t_{1} \ \ \ \ \ \ \ \ t_{2} \ \dotsc \ \ \ \ \ \ \ \ \ \ \ \ \ \ \ \ \ \ \ \ \ \ \ \ \ \ \ \ \ \ \ \ \ \ \ \ \ \ \ \ \ \ \ \ \ \ \ \ \ \ \ t_{d} \ \ \ \ t_{d+1} \ \ \ \ \ t_{d+2} \ \dotsc \ \ t_{d+m}$};
\draw (168.23,172.67) node  [font=\large] [align=left] {point process \\set predictor};
\draw (7.6,199.89) node [anchor=north west][inner sep=0.75pt]   [align=left] {desired risk $\displaystyle \alpha $};
\draw (99.53,16) node [anchor=north west][inner sep=0.75pt]   [align=left] {past events $\displaystyle x\in \mathbb{R}^{d}$};
\draw (273.14,151.91) node   [align=left] {predicted set\\ $\displaystyle \Gamma ( x|\mathcal{D}) \subseteq \mathbb{R}^{m}$};
\draw (498.67,53.33) node [anchor=north west][inner sep=0.75pt]   [align=left] {\begin{minipage}[lt]{21.99pt}\setlength\topsep{0pt}
\begin{flushright}
time
\end{flushright}

\end{minipage}};
\draw (337.33,150.73) node [anchor=north west][inner sep=0.75pt]    {$\Gamma _{1}$};
\draw (79.33,147.4) node [anchor=north west][inner sep=0.75pt]    {$\mathcal{D}$};
\draw (465.33,233.73) node [anchor=north west][inner sep=0.75pt]    {$\Gamma _{m}$};
\draw (382.67,170.4) node [anchor=north west][inner sep=0.75pt]    {$\dotsc $};
\draw (338.2,1) node [anchor=north west][inner sep=0.75pt]   [align=left] {future events $\displaystyle y\in \mathbb{R}^{m}$};

\end{tikzpicture}